\documentclass[11pt]{article}

\usepackage[preprint]{acl}

\usepackage{times}
\usepackage{latexsym}
\usepackage[T1]{fontenc}
\usepackage[utf8]{inputenc}
\usepackage{microtype}
\usepackage{inconsolata}
\usepackage{graphicx}

\usepackage{amsmath}
\usepackage{amssymb}
\usepackage{booktabs}
\usepackage{multirow}
\usepackage{xcolor}
\usepackage{colortbl}
\usepackage{nicematrix}
\usepackage{xspace}
\usepackage{listings}
\usepackage[most]{tcolorbox}
\usepackage{pifont}
\usepackage{tikz}
\usetikzlibrary{arrows.meta,positioning,fit,backgrounds,calc,shadows}

\newcommand{\LongZero}{\textsc{LongZero}\xspace}

\definecolor{avgbg}{HTML}{E3F1E4}
\newcommand{\avg}[1]{#1}%
\newcommand{\fst}[1]{\textbf{#1}}
\newcommand{\snd}[1]{\underline{#1}}
\definecolor{pbg}{HTML}{E8F0FB}
\definecolor{pframe}{HTML}{A9C4EB}
\newtcblisting{promptbox}[1]{enhanced, listing only, before skip=6pt, after skip=6pt,
  colback=pbg!45, colframe=pframe, boxrule=0.5pt, arc=2pt,
  left=4pt, right=4pt, top=3pt, bottom=3pt,
  fonttitle=\bfseries\footnotesize, coltitle=black, colbacktitle=pbg, title={#1},
  listing options={basicstyle=\fontfamily{zi4}\selectfont\footnotesize, breaklines=true, breakatwhitespace=true, breakindent=0pt, aboveskip=0pt, belowskip=0pt,
    columns=fullflexible, keepspaces=true, showstringspaces=false,
    literate={—}{{---}}1 {’}{{'}}1 {‘}{{`}}1 {“}{{``}}1 {”}{{''}}1 {±}{{$\pm$}}1}}

\makeatletter
\AtBeginDocument{%
  \let\lz@endthebib\endthebibliography
  \def\endthebibliography{\ifnum\c@NAT@ctr=\z@\item[]\fi\lz@endthebib}}
\makeatother

\title{\LongZero: Enhancing Long-Context Reasoning in LLMs from Zero Data}

\author{Zhennan Wan, Jianfei Chen\thanks{Corresponding Author} \\
  Dept. of Comp. Sci. and Tech., Institute for AI, 
  BNRist Center, THBI Lab, \\Tsinghua-Bosch Joint ML Center, 
  Tsinghua University\\
  \texttt{wanzn26@mails.tsinghua.edu.cn}, \texttt{jianfeic@tsinghua.edu.cn}}
\begin{document}
\maketitle

\begin{abstract}
Reinforcement learning (RL) is an effective way to strengthen the long-context reasoning of large language models (LLMs), but it needs training data that are long, diverse, challenging, and verifiable.
Human annotation of such data is expensive, naturally occurring data are too scarce to sustain RL, and existing synthesis methods depend on external corpora or stronger teacher models, which become harder to obtain as the trained models get stronger.
Recently, RL from zero data, in which a model learns from tasks it proposes itself, has achieved excellent performance, but existing methods target short-context tasks.
Therefore, we propose \textbf{\LongZero}, a method that lets an LLM improve its own long-context reasoning without any external data. 
Instead of writing questions and answers over existing documents, \LongZero first has the model synthesize the question and answer with its supporting evidence in a compact structured form, and only then separately expands the evidence into a long multi-document context.
\LongZero combines these self-written multi-hop reasoning tasks with self-written multi-round retrieval tasks and trains the model with GRPO.
We evaluate \LongZero on two LLMs across five long-context reasoning benchmarks and compare it with RL on data synthesized from external corpora. 
Experimental results show that \LongZero generalizes to real tasks and outperforms both baselines, suggesting a viable and promising path to improving long-context reasoning of LLMs.\footnote{Code and data are available at \url{https://github.com/zhennan1/LongZero}.}
\end{abstract}

\section{Introduction}
\label{sec:intro}

\begin{figure}[t]
\centering
\includegraphics[width=\columnwidth]{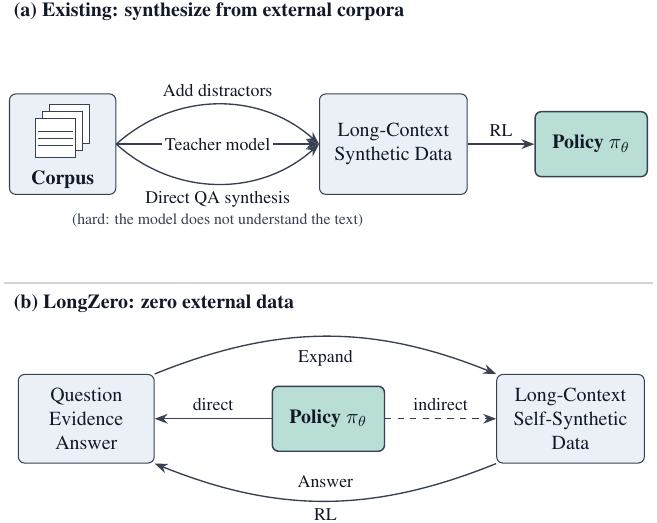}
\caption{Core idea of \LongZero. (a) Existing methods build long-context RL data from external corpora. (b) \LongZero lets the policy write the question, evidence, and answer, expand them into long-context data, and learn with RL from answering its own questions.}
\label{fig:intro}
\end{figure}

Reasoning over long context, such as collections of reports, long conversations, or entire books, is required in many applications of large language models (LLMs) \citep{bai2025longbench2,krishna2025frames}.
Following the success of reinforcement learning with verifiable rewards (RLVR) on short-context reasoning \citep{guo2025deepseekr1,kimi2025k15,lambert2024tulu3,yu2025dapo}, a growing body of work applies RL to long-context reasoning \citep{wan2025qwenlongl1,wang2025loongrl,shen2025qwenlongl15,chen2026longrlvr}.
These studies consistently show that \textbf{training data} are a decisive factor: useful RL data demand (i) substantial length, (ii) evidence retrieval and consolidation, (iii) sufficient difficulty for producing a learning signal, and (iv) automatically verifiable answers \citep{wang2025loongrl,xu2026beyond,lv2026golongrl}.

Data with all of these properties are difficult to obtain.
Human annotation of long-context questions is expensive, and naturally occurring long-context QA data are far scarcer than what RL consumes.
Existing pipelines therefore build training tasks from existing resources (Figure~\ref{fig:intro}a): they extend short multi-hop QA datasets with distractor documents \citep{wang2025loongrl}, ask a model to write questions over real documents \citep{yang2025spell,shen2025qwenlongl15,lv2026golongrl}, combine knowledge graphs with search-agent trajectories \citep{lin2026longtracerl}, or simply rely on a stronger teacher model to generate training data \citep{mukherjee2023orca,liu2024bestpractices}.
All of these approaches depend on external data or models. Real-world facts also introduce a confound: a model may answer from its parametric knowledge instead of the given context \citep{longpre2021entity}, and public text may overlap with evaluation data \citep{sainz2023nlp}.

At the same time, RL from \textbf{zero data} has emerged as an alternative to curated training sets.
In these methods the model proposes its own tasks and learns from solving them \citep{zhao2025absolute,huang2025rzero,chen2025sqlm,liu2025spiral}.
This line of work has achieved excellent performance in short-context domains, such as mathematics, code, games, and tool-using agents \citep{xia2025agent0,yue2026drzero,acikgoz2026toolr0}.
This leads to our research question: \textbf{Can an LLM also improve its long-context reasoning from zero data?}

Long-context tasks are harder to synthesize from zero data: a task consists of a long input together with a question about it, and the model must produce both.
Letting the model write a long document and then pose a question about it inherits the very weakness the task is meant to address, since a model that cannot yet reason reliably over long contexts cannot reliably write or verify questions about the long text it has just written.
Our key observation is that the order of generation can be reversed.
If the model first writes the \textbf{question, evidence, and answer} as a short structured specification, and only then expands the evidence into a long multi-document context, the answer is fixed before any long text exists, and the only remaining risk of error lies in the expansion.

Building on this observation, we propose \textbf{\LongZero} (Figure~\ref{fig:intro}b).
\LongZero uses the model being trained to synthesize two complementary kinds of long-context tasks: \textbf{reasoning} tasks, in which multi-hop questions over a fictional world must be answered from eight model-written documents, and \textbf{retrieval} tasks in the style of multi-round co-reference resolution (MRCR; \citealp{vodrahalli2024michelangelo}), in which the model must reproduce one specific earlier response in a long, model-written conversation.
The same model then answers each reasoning question several times, and only the questions it sometimes answers correctly and sometimes incorrectly are kept, because these are the questions on which group-relative policy optimization (GRPO; \citealp{shao2024deepseekmath}) receives a non-zero learning signal.

Our main contributions are as follows:
\begin{itemize}
  \item We propose \LongZero, a long-context RL method in which the LLM itself first synthesizes the question, evidence, and answer in a structured format, then expands them into long-context data, and is finally trained on these data.
  \item We evaluate \LongZero on two base models and five long-context reasoning benchmarks, and experimental results show that \LongZero can generalize from self-generated tasks to general long-context reasoning tasks.
  \item We compare \LongZero with two synthetic-data baselines built from external corpora, and experimental results show that \LongZero outperforms both baselines, demonstrating the effectiveness of \LongZero in improving long-context reasoning of LLMs without external data.
\end{itemize}

\section{Related Work}
\label{sec:related}

\paragraph{Reinforcement Learning from Zero Data.}
Learning without human data dates back to self-play systems such as AlphaGo Zero \citep{silver2017alphagozero}, and LLMs have long been used to generate their own instructions, rationales, or rewards \citep{wang2023selfinstruct,zelikman2022star,yuan2024selfrewarding}.
Since RLVR became the standard recipe for reasoning \citep{guo2025deepseekr1}, two lines of work have reduced its reliance on curated data.
The first removes labels: rewards are derived from majority voting \citep{zuo2025ttrl} or from the model's own confidence \citep{zhao2025intuitor}.
The second removes the questions themselves: the model proposes tasks and learns from solving them, verified by code execution in Absolute Zero \citep{zhao2025absolute}, by a co-evolving challenger and solver in R-Zero \citep{huang2025rzero}, or by asymmetric self-play from a single topic prompt \citep{chen2025sqlm}, and the idea has been extended to games \citep{liu2025spiral} and agents \citep{xia2025agent0,yue2026drzero}.
Programmatic task generators play a similar role for logic puzzles \citep{xie2025logicrl,stojanovski2025reasoninggym}.
All of these methods target short-context tasks, where a task is a short string; in long-context reasoning, the long context must itself be synthesized and kept consistent with the question and answer, which is the problem \LongZero addresses.

\paragraph{Long-Context Reinforcement Learning.}
Long-context ability is usually built by context extension and continued pre-training \citep{peng2024yarn,fu2024data,gao2025prolong}, followed by fine-tuning on synthetic long-context data \citep{bai2024longalign,an2024film,xiong2025artificial}.
RL has recently been applied directly to long-context reasoning, and most work focuses on how to construct training tasks: QwenLong-L1 and QwenLong-L1.5 write questions over real documents \citep{wan2025qwenlongl1,shen2025qwenlongl15}, LoongRL hides HotpotQA questions behind UUID chains among distractors \citep{wang2025loongrl}, SPELL lets one model question, answer, and verify over raw documents \citep{yang2025spell}, LongTraceRL builds questions from knowledge-graph walks and search trajectories \citep{lin2026longtracerl}, and GoLongRL curates a multi-task mixture \citep{lv2026golongrl}.
Other work improves the reward, for example with verifiable context rewards \citep{chen2026longrlvr}, or processes long inputs with memory agents \citep{yu2025memagent}.
These methods all obtain their contexts or questions from existing corpora, QA datasets, knowledge bases, or larger models; \LongZero synthesizes both with the model being trained and is complementary to better rewards and algorithms.

\paragraph{Evaluation of Long-Context Reasoning.}
Early evaluations relied on synthetic retrieval probes such as RULER \citep{hsieh2024ruler}, which revealed, for instance, that models miss information in the middle of long inputs \citep{liu2024lost}. Benchmarks have since moved toward realistic tasks and reasoning, such as LongBench and LongBench v2 \citep{bai2024longbench,bai2025longbench2}, MRCR \citep{vodrahalli2024michelangelo}, LongReason \citep{ling2025longreason}, and AA-LCR \citep{aa2025lcr}.

\section{\LongZero}
\label{sec:method}

\begin{figure*}[t]
\centering
\includegraphics[width=\textwidth]{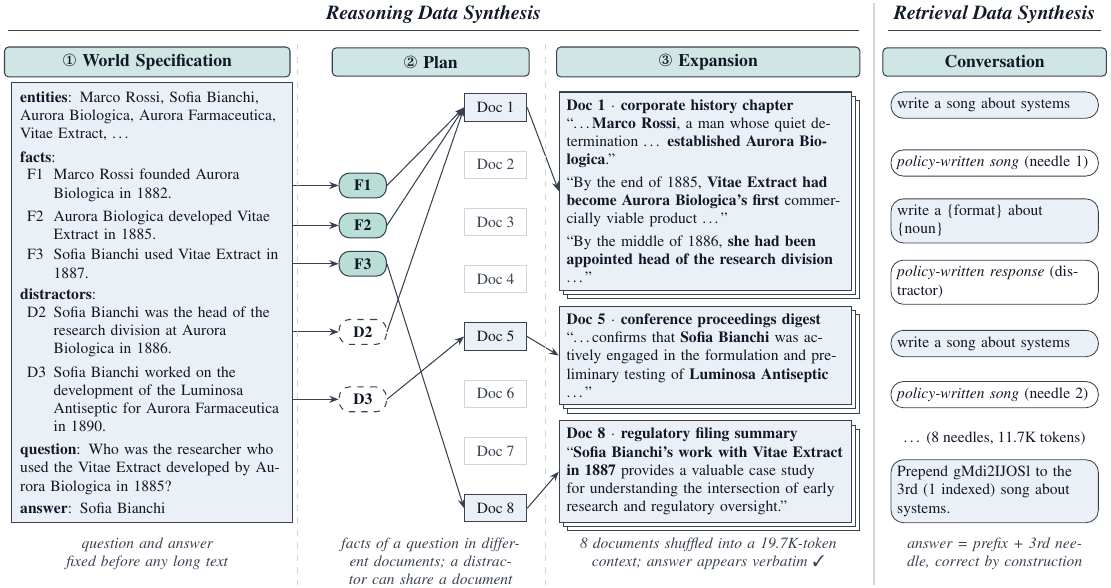}
\caption{Overview of \LongZero data generation, with real data synthesized by Qwen3-4B-Thinking-2507. Left: for reasoning data, the policy writes a world specification, plans which document states each fact, and expands the plan into long documents. Right: retrieval data are long conversations assembled from the policy's own responses.}
\label{fig:overview}
\end{figure*}

Let $\pi_\theta$ denote the LLM to be trained.
\LongZero uses $\pi_\theta$ itself to synthesize a training set $\mathcal{D}=\mathcal{D}_{\text{rea}}\cup\mathcal{D}_{\text{ret}}$ of long-context tasks with verifiable answers, and then optimizes $\pi_\theta$ on $\mathcal{D}$ with GRPO (Figure~\ref{fig:overview}).

\subsection{Data Synthesis}
\label{sec:synthesis}

\LongZero synthesizes two kinds of data: \textbf{reasoning data} $\mathcal{D}_{\text{rea}}$, which require finding facts scattered across long documents, rejecting similar-looking distractors, and combining the facts into an answer, and \textbf{retrieval data} $\mathcal{D}_{\text{ret}}$, which require locating and reproducing one specific item among many similar ones in a long conversation.
In both cases, the answer is determined by construction rather than by a model's judgment of a long text.

\subsubsection{Reasoning Data Synthesis}
\label{sec:reasoning}

A reasoning sample is drawn from a \textbf{world}: a set of fictional entities and facts rendered as $M$ long documents of different genres, together with multi-hop questions that require combining facts from different documents.
The difficulty of a question $q$ over a context $x$ grows with the length of the context, the number of hops, and the number of distractors:
\begin{equation}
\operatorname{Difficulty}(x,q)\propto |x|\cdot H_q\cdot K_q,
\end{equation}
where $|x|$ is the number of tokens in $x$, $H_q$ is the number of hops of $q$, and $K_q$ is the number of distractor facts that imitate its supporting facts.

We use this relation as a design guide rather than as a fitted model: the specification controls $H_q$ and $K_q$, and the expansion controls $|x|$.
The central design choice is to generate the world \textbf{top-down}: $\pi_\theta$ first writes a compact specification that fixes the questions, their supporting facts, and their answers, and only then expands it into long text.
All entities are fictional, which reduces reliance on parametric knowledge; fictional biographies are likewise used to study knowledge storage and extraction \citep{allenzhu2023physics31}.
A program first samples a domain, a set of genres, and a \textbf{style card} that fixes the time period, naming style, and banned names, so that worlds are diverse. The remaining steps are shown on the left of Figure~\ref{fig:overview}, and their prompts are given in Appendix~\ref{sec:prompts}.

\paragraph{Specification.}
$\pi_\theta$ writes a world specification $\mathcal{S}=(\mathcal{V},\mathcal{F},\mathcal{F}^{-},\mathcal{Q})$ as strict JSON.
$\mathcal{V}$ contains named entities, among which at least two people share a surname and at least two organizations have confusably similar names.
$\mathcal{F}$ contains atomic facts forming $Q$ question chains that partially share bridge entities, where the chain of question $q$ has $H_q$ hops, and $\mathcal{F}^{-}$ contains $K_q$ distractor facts per chain that involve the same bridge entity but differ in role, time, or attribute.
Each question $q\in\mathcal{Q}$ comes with its answer $y^\ast_q$ and supporting facts $F_q\subseteq\mathcal{F}$; it refers to intermediate entities by description, such as ``the founder of the magazine's parent press'', so that answering requires resolving each hop, whereas the answer must be a concrete name, place, product, title, or number appearing verbatim in a fact.
Because the specification is short, it can be generated and checked reliably, and it fixes the answer before any long text is written.

\paragraph{Plan and Expansion.}
$\pi_\theta$ assigns every fact in $\mathcal{F}\cup\mathcal{F}^{-}$ to one of the $M$ documents, spreading the supporting facts of each question across different documents and placing each distractor away from the fact it imitates, so that answering requires locating evidence in several documents and rejecting nearby distractors.
It then writes each document independently in the planned genre, stating each assigned fact exactly once and not stating facts assigned to other documents.

Finally, a program shuffles the $M$ documents into the context $x$ and keeps a question only if its answer or an equivalent alias appears verbatim in the expanded text; this faithfulness gate removes questions whose answer-bearing fact was dropped or altered during expansion.
Each surviving question yields a sample $(x,q,y^\ast_q)$.
  
\subsubsection{Retrieval Data Synthesis}
\label{sec:retrieval}

Our retrieval data follow the task format of MRCR \citep{vodrahalli2024michelangelo,openai2025mrcr}: in a long conversation, the user repeatedly asks for short pieces of the same kind, such as ``write a poem about tigers'', and finally asks the model to reproduce the $k$-th of them, prefixed by a random string $s$.
Instead of using existing conversations, \LongZero lets $\pi_\theta$ write all conversational content itself, as shown on the right of Figure~\ref{fig:overview}.
Given only a small set of formats and common nouns taken from the public OpenAI MRCR dataset \citep{openai2025mrcr}, $\pi_\theta$ responds to every request ``\texttt{write a \{format\} about \{noun\}}'' $R$ times, yielding a pool of self-written responses.
A program then assembles each sample from the pool: it selects a target request, inserts $N$ of $\pi_\theta$'s responses to that request, called \textbf{needles} $a_1,\dots,a_N$, at random positions among responses to other requests until the conversation reaches a length drawn from a fixed range, and asks for the $k$-th needle prefixed by $s$.

Because the program decides which response is the $k$-th needle, the reference answer $y^\ast=s\,\|\,a_k$ is correct by construction, while solving the task requires distinguishing the model's own near-duplicate responses and tracking their order over a long context.
We sample 1{,}152 training and 128 validation conversations without filtering.

\subsubsection{Difficulty Filtering}
\label{sec:filter}

\begin{table}[t]
\centering
\setlength{\tabcolsep}{4pt}
\resizebox{\columnwidth}{!}{%
\begin{tabular}{lrr}
\toprule
 & \textbf{Qwen3-4B-T} & \textbf{Qwen3-8B} \\
\midrule
\multicolumn{3}{l}{\textbf{Synthesis per 2{,}560 unstable reasoning questions}} \\
Worlds attempted & \textasciitilde{}3.8K & \textasciitilde{}5.4K \\
Worlds accepted & \textasciitilde{}3.6K & \textasciitilde{}5.1K \\
Questions filtered & \textasciitilde{}14.4K & \textasciitilde{}19.5K \\
\quad always correct & 69.3\% & 68.3\% \\
\quad unstable (kept) & 17.7\% & 13.1\% \\
\quad always incorrect & 13.0\% & 18.6\% \\
Filtering answers & \textasciitilde{}55K & \textasciitilde{}75K \\
\midrule
\multicolumn{3}{l}{\textbf{Training split (validation split)}} \\
Reasoning questions & 2{,}304 (256) & 2{,}304 (256) \\
Reasoning worlds & 1{,}705 (184) & 1{,}885 (206) \\
Reasoning tokens, median & 19.4K & 14.6K \\
Retrieval conversations & 1{,}152 (128) & 1{,}152 (128) \\
Retrieval tokens, median & 11.4K & 11.1K \\
\bottomrule
\end{tabular}%
}
\caption{Data synthesized by each model for itself. Upper: synthesis needed for the 2{,}560 unstable reasoning questions used, estimated from each model's rates. Lower: training (validation) data; token counts use each model's tokenizer. Qwen3-4B-T: Qwen3-4B-Thinking-2507.}
\label{tab:data}
\end{table}

GRPO normalizes rewards within a group of responses to the same prompt, so a question that the policy always answers correctly, or always answers incorrectly, contributes no gradient.
In our inspection, consistently failed questions were also frequently defective, for example because two documents assigned conflicting values to the same attribute.

\LongZero therefore lets $\pi_\theta$ measure the difficulty of its own questions and keeps only the \textbf{unstable band}.
Concretely, $\pi_\theta$ answers each question with the full context in small waves of samples, scored by the same matching rule as the training reward (\S\ref{sec:rl}); a question is kept as soon as it has received both a correct and an incorrect answer, and discarded if its first $W$ answers agree.
Because the same model generates and filters the data, the difficulty of the kept questions adapts to the model that will be trained on them.

From the unstable questions we hold out a validation set by world and form the training set from 2{,}304 reasoning questions and 1{,}152 retrieval conversations, a 2:1 mixture. As Table~\ref{tab:data} shows, Qwen3-8B keeps a smaller share of its own questions than Qwen3-4B-Thinking-2507, 13.1\% versus 17.7\%, and therefore needs more synthesis to obtain the same amount of data.

\begin{table*}[t]
\centering
\setlength{\tabcolsep}{8pt}
\resizebox{\textwidth}{!}{%
\begin{NiceTabular}{lcccccc}
\CodeBefore
\columncolor{avgbg}{7}
\Body
\toprule
\textbf{Method} & \textbf{LB1} & \textbf{LBv2} & \textbf{FRAMES} & \textbf{AA-LCR} & \textbf{LongReason} & \avg{\textbf{Avg}} \\
\midrule
\multicolumn{6}{l}{\textbf{Qwen3-4B-Thinking-2507}} & \avg{} \\
Base model & 65.43 & 40.56 & 51.27 & 36.50 & 79.66 & \avg{54.68} \\
Enhanced HotpotQA (from LoongRL) & 65.94 & \snd{43.04} & 51.46 & \fst{41.50} & \snd{80.92} & \avg{\snd{56.57}} \\
Teacher-synthesized QA & \snd{66.12} & 41.85 & \snd{52.97} & 38.50 & 80.48 & \avg{55.99} \\
\LongZero (Ours) & \fst{67.22} & \fst{43.34} & \fst{55.76} & \snd{39.50} & \fst{81.30} & \avg{\fst{57.42}} \\
\midrule
\multicolumn{6}{l}{\textbf{Qwen3-8B (Thinking)}} & \avg{} \\
Base model & 63.63 & \snd{32.11} & \snd{48.00} & \snd{2.50} & 76.89 & \avg{44.63} \\
Enhanced HotpotQA (from LoongRL) & \snd{64.98} & 31.31 & 47.88 & \fst{3.00} & \snd{77.96} & \avg{\snd{45.03}} \\
Teacher-synthesized QA & 64.07 & \fst{33.20} & 47.94 & 1.00 & \snd{77.96} & \avg{44.83} \\
\LongZero (Ours) & \fst{65.49} & 32.01 & \fst{49.39} & \fst{3.00} & \fst{78.40} & \avg{\fst{45.66}} \\
\bottomrule
\end{NiceTabular}%
}
\caption{Results on five long-context reasoning benchmarks (mean of two runs; LongReason: 32K split; Avg: unweighted mean). Best results for each base model are in \textbf{bold} and second-best results are \underline{underlined}. All methods share the RL configuration and checkpoint step.}
\label{tab:main}
\end{table*}

\subsection{Reinforcement Learning}
\label{sec:rl}

We optimize $\pi_\theta$ with Group Relative Policy Optimization (GRPO; \citealp{shao2024deepseekmath}), which removes the value model of PPO \citep{schulman2017ppo} and instead uses the average reward of several outputs to the same prompt as the baseline.
For each prompt $x$, GRPO samples a group of $G$ outputs $\{o_1,\dots,o_G\}$ from the old policy $\pi_{\theta_{\text{old}}}$ and scores them with the rule-based rewards below, giving $\mathbf{r}=\{r_1,\dots,r_G\}$.
With outcome rewards, all tokens of $o_i$ share the group-normalized reward as their advantage:
\begin{equation}
\hat{A}_{i,t}=\frac{r_i-\operatorname{mean}(\mathbf{r})}{\operatorname{std}(\mathbf{r})}.
\end{equation}
Following DAPO \citep{yu2025dapo}, we use a token-level policy-gradient loss, which averages over all tokens in the group rather than first within each output, and maximize
\begin{align}
\mathcal{J}(\theta)&=\mathbb{E}_{x,\{o_i\}}\Big[\tfrac{1}{\sum_{i}|o_i|}\textstyle\sum_{i=1}^{G}\sum_{t=1}^{|o_i|}\ell_{i,t}(\theta)\Big],\\
\ell_{i,t}(\theta)&=\min\big(\rho_{i,t}(\theta)\hat{A}_{i,t},\notag\\
&\qquad\operatorname{clip}(\rho_{i,t}(\theta),1-\varepsilon,1+\varepsilon)\hat{A}_{i,t}\big)\notag\\
&\quad-\beta\,\hat{\mathbb{D}}_{\text{KL}}\big[\pi_\theta\,\|\,\pi_{\text{ref}}\big],
\end{align}
where the expectation is over prompts $x$ and groups $\{o_i\}_{i=1}^{G}\sim\pi_{\theta_{\text{old}}}(\cdot\mid x)$, $\rho_{i,t}(\theta)=\pi_\theta(o_{i,t}\mid x,o_{i,<t})/\pi_{\theta_{\text{old}}}(o_{i,t}\mid x,o_{i,<t})$ is the importance ratio, $\varepsilon$ is the clipping range, and $\pi_{\text{ref}}$ is the initial policy.
As in GRPO, the KL penalty with coefficient $\beta$ is added to the loss rather than to the reward, and is estimated per token with the k3 estimator of \citet{schulman2020kl}:
\begin{equation}
\hat{\mathbb{D}}_{\text{KL}}\big[\pi_\theta\,\|\,\pi_{\text{ref}}\big]=\kappa_{i,t}-\log\kappa_{i,t}-1,
\end{equation}
where $\kappa_{i,t}=\pi_{\text{ref}}(o_{i,t}\mid x,o_{i,<t})/\pi_\theta(o_{i,t}\mid x,o_{i,<t})$.
Both the token-level loss and this KL estimator follow the GRPO implementation of verl \citep{sheng2025hybridflow}, which we use for training.

\paragraph{Reward Design.}
Rewards are rule-based and selected by the data source of each sample.
For \textbf{reasoning} data, the prompt asks the model to end with ``\texttt{The answer is <answer>.}''
Let $\hat{y}$ be the text after the last ``the answer is'' in the output. The reward is
\begin{equation}
r^{\text{rea}}=\begin{cases}
1, & \text{if } \hat{y} \text{ matches a reference answer},\\
0, & \text{otherwise}.
\end{cases}
\end{equation}
The reference answers are the answer $y^\ast_q$ and its aliases. $\hat{y}$ matches a reference answer if it contains that answer as a whole-word span, so that ``15'' does not match ``5'', and is at most $L$ characters longer than it, which prevents rewarding answers that list many candidates.
Before matching, both texts are lowercased, quotation marks are unified, punctuation is replaced by spaces, and whitespace is collapsed.

For \textbf{retrieval} data, the reward is 0 if the final answer after the thinking part does not start with the prefix $s$.
Otherwise, it is the similarity between the rest of the answer and the needle $a_k$, measured by the \texttt{SequenceMatcher} ratio of Python's \texttt{difflib} as in the MRCR evaluation \citep{openai2025mrcr}; this dense reward gives partial credit for nearly correct reproductions.

\section{Experiments}
\label{sec:experiments}
\begin{figure*}[t]
\centering
\includegraphics[width=0.8\textwidth]{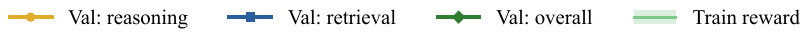}\\[2pt]
\setlength{\tabcolsep}{2pt}
\begin{tabular}{cccc}
 & \small\textbf{Reasoning} & \small\textbf{Retrieval} & \small\textbf{Overall} \\
\rotatebox{90}{\small\hspace{1mm}Qwen3-4B-Thinking-2507} &
\includegraphics[width=0.305\textwidth]{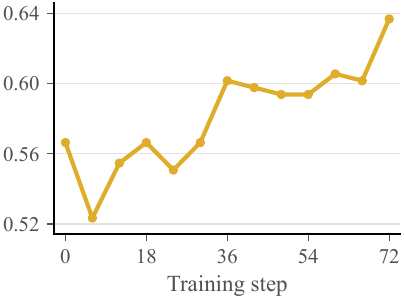} &
\includegraphics[width=0.305\textwidth]{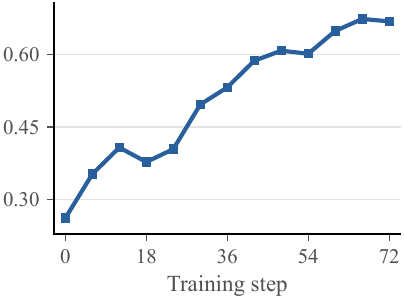} &
\includegraphics[width=0.305\textwidth]{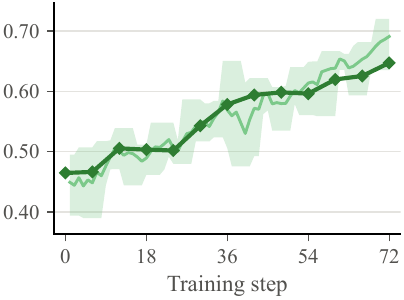} \\
\rotatebox{90}{\small\hspace{9mm}Qwen3-8B} &
\includegraphics[width=0.305\textwidth]{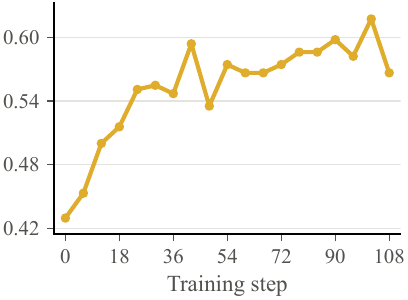} &
\includegraphics[width=0.305\textwidth]{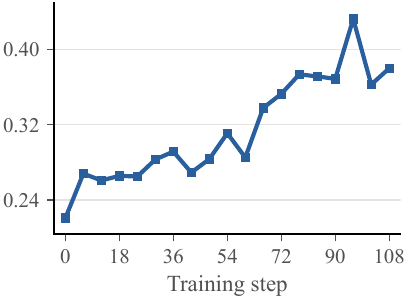} &
\includegraphics[width=0.305\textwidth]{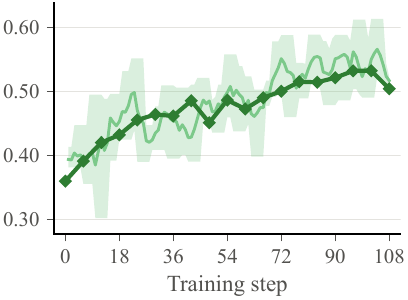} \\
\end{tabular}
\caption{Training dynamics of \LongZero up to the reported checkpoints. Validation scores are measured every 6 steps on held-out reasoning and retrieval data (overall: weighted by set size). The training reward mixes both data types (line: 5-step average; band: range over the same window).}
\label{fig:curves}
\end{figure*}

\subsection{Settings}
\label{sec:settings}

\paragraph{Training Setup.}
We use (i) \textbf{Qwen3-4B-Thinking-2507} and (ii) \textbf{Qwen3-8B} in thinking mode \citep{yang2025qwen3}.
Each model synthesizes and filters its own data (Table~\ref{tab:data}) and is trained with GRPO in verl \citep{sheng2025hybridflow} for 72 steps on Qwen3-4B-Thinking-2507 and 108 steps on Qwen3-8B, about 1.3 and 2 epochs of our data, with 64 prompts and $G{=}8$ responses per step, a constant learning rate of $10^{-6}$, $\varepsilon{=}0.2$, $\beta{=}0.001$, and at most 32{,}768 prompt and 8{,}192 response tokens.
All methods, including the baselines, share this configuration, and we report the final checkpoint of each run; Appendix~\ref{sec:exp-details} lists all hyperparameters.

\paragraph{Benchmarks.}
We use five long-context reasoning benchmarks: (i) the multi-document QA subsets of \textbf{LongBench} \citep[LB1;][]{bai2024longbench}, namely HotpotQA, 2Wiki\-Multihop\-QA, and MuSiQue \citep{yang2018hotpotqa,ho2020constructing,trivedi2022musique} with F1; (ii) \textbf{LongBench v2} \citep[LBv2;][]{bai2025longbench2}, 503 four-choice questions; (iii) \textbf{FRAMES} \citep{krishna2025frames} with oracle Wikipedia articles, 824 questions scored by answer containment; (iv) \textbf{AA-LCR} \citep{aa2025lcr}, 100 questions over document sets of 72K--115K tokens, judged by GPT-5.6-Luna with the official judging prompt; and (v) the 32K split of \textbf{LongReason} \citep{ling2025longreason}, 794 multiple-choice questions.

We sample with temperature 0.7 and top-$p$ 0.95, since greedy decoding is not recommended for Qwen3 thinking models, run each benchmark at least twice, and report the mean of two runs.
Qwen3-4B-Thinking-2507 natively supports 262K tokens, but Qwen3-8B supports only 32K. Following the Qwen3-8B model card, we apply YaRN \citep{peng2024yarn} with a scaling factor of 4 at evaluation time, without additional training; RL training does not use YaRN.

\paragraph{Baselines.}
Besides the base models, we compare with two baselines that differ from \LongZero only in their training data:
(i) \textbf{Enhanced HotpotQA (from LoongRL)}: 5{,}120 difficulty-filtered HotpotQA \citep{yang2018hotpotqa} questions whose contexts are extended to 16K tokens with distracting documents by the data pipeline of LoongRL \citep{wang2025loongrl}, trained with LoongRL's prompt and exact-match reward;
(ii) \textbf{Teacher-synthesized QA}: multi-hop questions written and verified by Qwen3.5-397B-A17B over Ultra-FineWeb \citep{wang2025ultrafineweb} document clusters from SPELL \citep{yang2025spell}; each base model keeps 2{,}304 questions it answers correctly 1--7 times out of 8, mixed 2:1 with the same self-synthesized retrieval data and trained with the same prompt and reward, so that only the source of the reasoning questions differs. Appendix~\ref{sec:exp-details} describes both baselines in detail.

\subsection{Results and Analysis}
\label{sec:results}

\paragraph{Overall Performance.}
Table~\ref{tab:main} shows that \LongZero achieves the highest average score on both base models compared to the baselines. All three RL methods improve over the base models on average, but \LongZero gains the most and is best on four of the five benchmarks for both models, counting a tie on AA-LCR for Qwen3-8B; the exceptions are AA-LCR on Qwen3-4B-Thinking-2507 and LBv2 on Qwen3-8B. Notably, the teacher-synthesized baseline does not surpass ours, although its questions and answers are written and verified by a 397B-parameter model. A stronger generator thus does not guarantee better training data: questions written by another model over real documents need not match the ability of the policy, whereas those of \LongZero are written by the policy itself, verifiable by construction, and filtered to its current ability.

\begin{table*}[t]
\centering
\setlength{\tabcolsep}{8pt}
\resizebox{\textwidth}{!}{%
\begin{NiceTabular}{lcccccc}
\CodeBefore
\columncolor{avgbg}{7}
\Body
\toprule
\textbf{Training data} & \textbf{LB1} & \textbf{LBv2} & \textbf{FRAMES} & \textbf{AA-LCR} & \textbf{LongReason} & \avg{\textbf{Avg}} \\
\midrule
Base model & 65.43 & 40.56 & 51.27 & \snd{36.50} & 79.66 & \avg{54.68} \\
Reasoning data only & 67.04 & \snd{42.64} & \snd{54.61} & 34.00 & 78.59 & \avg{55.38} \\
Retrieval data only & 64.51 & 41.65 & 52.18 & 33.00 & 77.90 & \avg{53.85} \\
Reasoning and retrieval data, 1:1 & \fst{67.29} & 41.35 & 53.64 & \snd{36.50} & \snd{80.98} & \avg{\snd{55.95}} \\
Reasoning and retrieval data, 2:1 (Ours) & \snd{67.22} & \fst{43.34} & \fst{55.76} & \fst{39.50} & \fst{81.30} & \avg{\fst{57.42}} \\
\bottomrule
\end{NiceTabular}%
}
\caption{Ablation on the training data of Qwen3-4B-Thinking-2507: reasoning data only, retrieval data only, and mixtures of the two in a 1:1 and a 2:1 ratio. All runs share the RL configuration. Best results are in \textbf{bold} and second-best results are \underline{underlined}.}
\label{tab:ablation}
\end{table*}

\paragraph{Generalization across Tasks.}
Figure~\ref{fig:curves} shows that validation scores on held-out synthesized data rise over training for both data types and both models, with fluctuations but no early saturation. More importantly, these in-distribution gains transfer to the general long-context reasoning benchmarks, which do not share the entities of the training data. On Qwen3-4B-Thinking-2507, \LongZero improves on every benchmark, including LB1 and FRAMES, whose questions concern real Wikipedia entities, although every training world is fictional. Because questions about fictional entities cannot be answered from memory, the policy can only improve on them by reading the context more carefully, and this skill carries over to real documents. The benchmark gains are smaller than the validation gains, as expected under distribution shift, but appear across extractive, multiple-choice, and model-judged free-form answers. The baselines generalize less evenly: they often improve on certain tasks while declining on others. On Qwen3-8B, each falls below the base model on two benchmarks, whereas \LongZero does so only on LBv2, and on Qwen3-4B-Thinking-2507, Enhanced HotpotQA does not improve on the HotpotQA subset of LB1 (Table~\ref{tab:lb1-4b}) despite training on HotpotQA.

\paragraph{Generalization across Models.}
\LongZero improves both base models even though each one synthesizes and filters its own data, so the method does not depend on a particular generator. The improvement is clearly larger on Qwen3-4B-Thinking-2507, which is also the stronger long-context reasoner of the two. This is consistent with the self-synthesis loop: a stronger model is likely to write more faithful specifications and documents, and it wastes less of its synthesis on questions it can never answer (Table~\ref{tab:data}), so its training data are both cleaner and better matched to its ability. Qwen3-8B does not improve on LBv2, and all methods score near zero on AA-LCR, whose 72K--115K-token inputs leave little room for output within the 131K-token window that YaRN provides, so the thinking model often runs out of tokens before giving a final answer: 120 of the 200 AA-LCR responses of \LongZero on Qwen3-8B contain no answer.

\paragraph{Length Generalization and Efficiency.}
All training contexts are short compared with the benchmarks: reasoning and retrieval samples contain tens of thousands of tokens at most, whereas LBv2 and AA-LCR require reading inputs that are several times longer. \LongZero nevertheless improves Qwen3-4B-Thinking-2507 on both of these benchmarks, so the skills learned on short synthetic contexts, namely locating scattered evidence and rejecting look-alike distractors, generalize to much longer inputs. Moreover, the whole pipeline runs on the model being trained: unlike the teacher-synthesized baseline, which needs a 397B-parameter model to write, answer, and verify every question, \LongZero requires no calls to a larger model, and its data cost scales with the size of the policy rather than with that of a teacher.

\paragraph{Data Complementarity.}
Table~\ref{tab:ablation} shows that the two kinds of synthesized data are complementary. Reasoning data alone already give most of the improvement on LB1 and FRAMES but fall below the base model on AA-LCR and LongReason. Retrieval data alone lower the average. Combining them is better on average than either alone, and the 2:1 mixture used by \LongZero is the best variant on four of the five benchmarks and improves over the base model on all of them, avoiding the losses that each data type causes on its own. We attribute this to the different skills the two tasks train: reasoning data teach the policy to chain facts across documents, while retrieval data teach it to locate and reproduce one exact item among many similar ones. The better result of the 2:1 mixture over the 1:1 mixture further suggests that reasoning data should dominate, with retrieval data acting as a basic signal.

\section{Conclusion and Future Work}
\label{sec:conclusion}

We presented \LongZero, a method for improving the long-context reasoning of LLMs from zero data.
\LongZero has the model being trained write a structured specification of the question, evidence, and answer before expanding it into long documents, so that answers are correct by construction without any external corpus, human annotation, or teacher model; it combines these reasoning tasks with self-synthesized retrieval tasks and keeps only the questions the model answers inconsistently.
On two base models and five long-context reasoning benchmarks, \LongZero achieves the highest average score and outperforms RL on Enhanced HotpotQA and on questions written by a much larger teacher model, although it uses no data beyond the model itself. Its gains transfer from fictional to real-world documents and from short training contexts to much longer inputs, and its two data types are complementary.
Future work includes iterating \LongZero so that the trained model regenerates and refilters its own data, synthesizing longer contexts by writing more documents per world, adding question types that require counting, comparison, or arithmetic over the specification, and exploring better ways to ensure the correctness of the data.

\bibliography{longzero_arxiv}

@misc{aa2025lcr,
  title = {{Artificial Analysis Long Context Reasoning (AA-LCR)}},
  author = {{Artificial Analysis}},
  howpublished = {\url{https://huggingface.co/datasets/ArtificialAnalysis/AA-LCR}},
  year = {2025}
}

@article{acikgoz2026toolr0,
  title = {{Tool-R0: Self-Evolving LLM Agents for Tool-Learning from Zero Data}},
  author = {Acikgoz, Emre Can and Qian, Cheng and Hübotter, Jonas and Ji, Heng and Hakkani-Tür, Dilek and Tur, Gokhan},
  journal = {arXiv preprint arXiv:2602.21320},
  year = {2026},
  url = {https://arxiv.org/abs/2602.21320}
}

@article{allenzhu2023physics31,
  title = {{Physics of Language Models: Part 3.1, Knowledge Storage and Extraction}},
  author = {Allen-Zhu, Zeyuan and Li, Yuanzhi},
  journal = {arXiv preprint arXiv:2309.14316},
  year = {2023},
  url = {https://arxiv.org/abs/2309.14316}
}

@article{an2024film,
  title = {{Make Your LLM Fully Utilize the Context}},
  author = {An, Shengnan and Ma, Zexiong and Lin, Zeqi and Zheng, Nanning and Lou, Jian-Guang},
  journal = {arXiv preprint arXiv:2404.16811},
  year = {2024},
  url = {https://arxiv.org/abs/2404.16811}
}

@article{bai2024longalign,
  title = {{LongAlign: A Recipe for Long Context Alignment of Large Language Models}},
  author = {Bai, Yushi and Lv, Xin and Zhang, Jiajie and He, Yuze and Qi, Ji and Hou, Lei and Tang, Jie and Dong, Yuxiao and Li, Juanzi},
  journal = {arXiv preprint arXiv:2401.18058},
  year = {2024},
  url = {https://arxiv.org/abs/2401.18058}
}

@inproceedings{bai2024longbench,
  title = {{LongBench: A Bilingual, Multitask Benchmark for Long Context Understanding}},
  author = {Bai, Yushi and Lv, Xin and Zhang, Jiajie and Lyu, Hongchang and Tang, Jiankai and Huang, Zhidian and Du, Zhengxiao and Liu, Xiao and Zeng, Aohan and Hou, Lei and Dong, Yuxiao and Tang, Jie and Li, Juanzi},
  booktitle = {Proceedings of the 62nd Annual Meeting of the Association for Computational Linguistics (Volume 1: Long Papers)},
  year = {2024},
  url = {https://arxiv.org/abs/2308.14508}
}

@inproceedings{bai2025longbench2,
  title = {{LongBench v2: Towards Deeper Understanding and Reasoning on Realistic Long-context Multitasks}},
  author = {Bai, Yushi and Tu, Shangqing and Zhang, Jiajie and Peng, Hao and Wang, Xiaozhi and Lv, Xin and Cao, Shulin and Xu, Jiazheng and Hou, Lei and Dong, Yuxiao and Tang, Jie and Li, Juanzi},
  booktitle = {Proceedings of the 63rd Annual Meeting of the Association for Computational Linguistics (Volume 1: Long Papers)},
  year = {2025},
  url = {https://arxiv.org/abs/2412.15204}
}

@article{chen2025sqlm,
  title = {{Self-Questioning Language Models}},
  author = {Chen, Lili and Prabhudesai, Mihir and Fragkiadaki, Katerina and Liu, Hao and Pathak, Deepak},
  journal = {arXiv preprint arXiv:2508.03682},
  year = {2025},
  url = {https://arxiv.org/abs/2508.03682}
}

@article{chen2026longrlvr,
  title = {{LongRLVR: Long-Context Reinforcement Learning Requires Verifiable Context Rewards}},
  author = {Chen, Guanzheng and Shieh, Michael Qizhe and Bing, Lidong},
  journal = {arXiv preprint arXiv:2603.02146},
  year = {2026},
  url = {https://arxiv.org/abs/2603.02146}
}

@inproceedings{fu2024data,
  title = {{Data Engineering for Scaling Language Models to 128K Context}},
  author = {Fu, Yao and Panda, Rameswar and Niu, Xinyao and Yue, Xiang and Hajishirzi, Hannaneh and Kim, Yoon and Peng, Hao},
  booktitle = {Proceedings of the 41st International Conference on Machine Learning},
  year = {2024},
  url = {https://arxiv.org/abs/2402.10171}
}

@article{gao2025prolong,
  title = {{How to Train Long-Context Language Models (Effectively)}},
  author = {Gao, Tianyu and Wettig, Alexander and Yen, Howard and Chen, Danqi},
  journal = {arXiv preprint arXiv:2410.02660},
  year = {2024},
  url = {https://arxiv.org/abs/2410.02660}
}

@article{guo2025deepseekr1,
  title = {{DeepSeek-R1: Incentivizing Reasoning Capability in LLMs via Reinforcement Learning}},
  author = {DeepSeek-AI and Guo, Daya and Yang, Dejian and Zhang, Haowei and Song, Junxiao and Wang, Peiyi and Zhu, Qihao and Xu, Runxin and Zhang, Ruoyu and Ma, Shirong and Bi, Xiao and Zhang, Xiaokang and Yu, Xingkai and Wu, Yu and Wu, Z. F. and Gou, Zhibin and Shao, Zhihong and Li, Zhuoshu and Gao, Ziyi and Liu, Aixin and Xue, Bing and Wang, Bingxuan and Wu, Bochao and Feng, Bei and Lu, Chengda and Zhao, Chenggang and Deng, Chengqi and Zhang, Chenyu and Ruan, Chong and Dai, Damai and Chen, Deli and Ji, Dongjie and Li, Erhang and Lin, Fangyun and Dai, Fucong and Luo, Fuli and Hao, Guangbo and Chen, Guanting and Li, Guowei and Zhang, H. and Bao, Han and Xu, Hanwei and Wang, Haocheng and Ding, Honghui and Xin, Huajian and Gao, Huazuo and Qu, Hui and Li, Hui and Guo, Jianzhong and Li, Jiashi and Wang, Jiawei and Chen, Jingchang and Yuan, Jingyang and Qiu, Junjie and Li, Junlong and Cai, J. L. and Ni, Jiaqi and Liang, Jian and Chen, Jin and Dong, Kai and Hu, Kai and Gao, Kaige and Guan, Kang and Huang, Kexin and Yu, Kuai and Wang, Lean and Zhang, Lecong and Zhao, Liang and Wang, Litong and Zhang, Liyue and Xu, Lei and Xia, Leyi and Zhang, Mingchuan and Zhang, Minghua and Tang, Minghui and Li, Meng and Wang, Miaojun and Li, Mingming and Tian, Ning and Huang, Panpan and Zhang, Peng and Wang, Qiancheng and Chen, Qinyu and Du, Qiushi and Ge, Ruiqi and Zhang, Ruisong and Pan, Ruizhe and Wang, Runji and Chen, R. J. and Jin, R. L. and Chen, Ruyi and Lu, Shanghao and Zhou, Shangyan and Chen, Shanhuang and Ye, Shengfeng and Wang, Shiyu and Yu, Shuiping and Zhou, Shunfeng and Pan, Shuting and Li, S. S. and Zhou, Shuang and Wu, Shaoqing and Ye, Shengfeng and Yun, Tao and Pei, Tian and Sun, Tianyu and Wang, T. and Zeng, Wangding and Zhao, Wanjia and Liu, Wen and Liang, Wenfeng and Gao, Wenjun and Yu, Wenqin and Zhang, Wentao and Xiao, W. L. and An, Wei and Liu, Xiaodong and Wang, Xiaohan and Chen, Xiaokang and Nie, Xiaotao and Cheng, Xin and Liu, Xin and Xie, Xin and Liu, Xingchao and Yang, Xinyu and Li, Xinyuan and Su, Xuecheng and Lin, Xuheng and Li, X. Q. and Jin, Xiangyue and Shen, Xiaojin and Chen, Xiaosha and Sun, Xiaowen and Wang, Xiaoxiang and Song, Xinnan and Zhou, Xinyi and Wang, Xianzu and Shan, Xinxia and Li, Y. K. and Wang, Y. Q. and Wei, Y. X. and Zhang, Yang and Xu, Yanhong and Li, Yao and Zhao, Yao and Sun, Yaofeng and Wang, Yaohui and Yu, Yi and Zhang, Yichao and Shi, Yifan and Xiong, Yiliang and He, Ying and Piao, Yishi and Wang, Yisong and Tan, Yixuan and Ma, Yiyang and Liu, Yiyuan and Guo, Yongqiang and Ou, Yuan and Wang, Yuduan and Gong, Yue and Zou, Yuheng and He, Yujia and Xiong, Yunfan and Luo, Yuxiang and You, Yuxiang and Liu, Yuxuan and Zhou, Yuyang and Zhu, Y. X. and Xu, Yanhong and Huang, Yanping and Li, Yaohui and Zheng, Yi and Zhu, Yuchen and Ma, Yunxian and Tang, Ying and Zha, Yukun and Yan, Yuting and Ren, Z. Z. and Ren, Zehui and Sha, Zhangli and Fu, Zhe and Xu, Zhean and Xie, Zhenda and Zhang, Zhengyan and Hao, Zhewen and Ma, Zhicheng and Yan, Zhigang and Wu, Zhiyu and Gu, Zihui and Zhu, Zijia and Liu, Zijun and Li, Zilin and Xie, Ziwei and Song, Ziyang and Pan, Zizheng and Huang, Zhen and Xu, Zhipeng and Zhang, Zhongyu and Zhang, Zhen},
  journal = {arXiv preprint arXiv:2501.12948},
  year = {2025},
  url = {https://arxiv.org/abs/2501.12948}
}

@inproceedings{ho2020constructing,
  title = {{Constructing A Multi-hop QA Dataset for Comprehensive Evaluation of Reasoning Steps}},
  author = {Ho, Xanh and Nguyen, Anh-Khoa Duong and Sugawara, Saku and Aizawa, Akiko},
  booktitle = {Proceedings of the 28th International Conference on Computational Linguistics},
  year = {2020},
  url = {https://arxiv.org/abs/2011.01060}
}

@inproceedings{hsieh2024ruler,
  title = {{RULER: What's the Real Context Size of Your Long-Context Language Models?}},
  author = {Hsieh, Cheng-Ping and Sun, Simeng and Kriman, Samuel and Acharya, Shantanu and Rekesh, Dima and Jia, Fei and Zhang, Yang and Ginsburg, Boris},
  booktitle = {First Conference on Language Modeling},
  year = {2024},
  url = {https://arxiv.org/abs/2404.06654}
}

@article{huang2025rzero,
  title = {{R-Zero: Self-Evolving Reasoning LLM from Zero Data}},
  author = {Huang, Chengsong and Yu, Wenhao and Wang, Xiaoyang and Zhang, Hongming and Li, Zongxia and Li, Ruosen and Huang, Jiaxin and Mi, Haitao and Yu, Dong},
  journal = {arXiv preprint arXiv:2508.05004},
  year = {2025},
  url = {https://arxiv.org/abs/2508.05004}
}

@article{jacobs2023ulysses,
  title = {{DeepSpeed Ulysses: System Optimizations for Enabling Training of Extreme Long Sequence Transformer Models}},
  author = {Jacobs, Sam Ade and Tanaka, Masahiro and Zhang, Chengming and Zhang, Minjia and Song, Shuaiwen Leon and Rajbhandari, Samyam and He, Yuxiong},
  journal = {arXiv preprint arXiv:2309.14509},
  year = {2023},
  url = {https://arxiv.org/abs/2309.14509}
}

@article{kimi2025k15,
  title = {{Kimi k1.5: Scaling Reinforcement Learning with LLMs}},
  author = {Kimi Team and Du, Angang and Gao, Bofei and Xing, Bowei and Jiang, Changjiu and Chen, Cheng and Li, Cheng and Xiao, Chenjun and Du, Chenzhuang and Liao, Chonghua and Tang, Chuning and Wang, Congcong and Zhang, Dehao and Yuan, Enming and Lu, Enzhe and Tang, Fengxiang and Sung, Flood and Wei, Guangda and Lai, Guokun and Guo, Haiqing and Zhu, Han and Ding, Hao and Hu, Hao and Yang, Hao and Zhang, Hao and Yao, Haotian and Zhao, Haotian and Lu, Haoyu and Li, Haoze and Yu, Haozhen and Gao, Hongcheng and Zheng, Huabin and Yuan, Huan and Chen, Jia and Guo, Jianhang and Su, Jianlin and Wang, Jianzhou and Zhao, Jie and Zhang, Jin and Liu, Jingyuan and Yan, Junjie and Wu, Junyan and Shi, Lidong and Ye, Ling and Yu, Longhui and Dong, Mengnan and Zhang, Neo and Ma, Ningchen and Pan, Qiwei and Gong, Qucheng and Liu, Shaowei and Ma, Shengling and Wei, Shupeng and Cao, Sihan and Huang, Siying and Jiang, Tao and Gao, Weihao and Xiong, Weimin and He, Weiran and Huang, Weixiao and Xu, Weixin and Wu, Wenhao and He, Wenyang and Wei, Xianghui and Jia, Xianqing and Wu, Xingzhe and Xu, Xinran and Zu, Xinxing and Zhou, Xinyu and Pan, Xuehai and Charles, Y. and Li, Yang and Hu, Yangyang and Liu, Yangyang and Chen, Yanru and Wang, Yejie and Liu, Yibo and Qin, Yidao and Liu, Yifeng and Yang, Ying and Bao, Yiping and Du, Yulun and Wu, Yuxin and Wang, Yuzhi and Zhou, Zaida and Wang, Zhaoji and Li, Zhaowei and Zhu, Zhen and Zhang, Zheng and Wang, Zhexu and Yang, Zhilin and Huang, Zhiqi and Huang, Zihao and Xu, Ziyao and Yang, Zonghan and Lin, Zongyu},
  journal = {arXiv preprint arXiv:2501.12599},
  year = {2025},
  url = {https://arxiv.org/abs/2501.12599}
}

@inproceedings{krishna2025frames,
  title = {{Fact, Fetch, and Reason: A Unified Evaluation of Retrieval-Augmented Generation}},
  author = {Krishna, Satyapriya and Krishna, Kalpesh and Mohananey, Anhad and Schwarcz, Steven and Stambler, Adam and Upadhyay, Shyam and Faruqui, Manaal},
  booktitle = {Proceedings of the 2025 Conference of the Nations of the Americas Chapter of the Association for Computational Linguistics: Human Language Technologies (Volume 1: Long Papers)},
  year = {2025},
  url = {https://arxiv.org/abs/2409.12941}
}

@inproceedings{kwon2023vllm,
  title = {{Efficient Memory Management for Large Language Model Serving with PagedAttention}},
  author = {Kwon, Woosuk and Li, Zhuohan and Zhuang, Siyuan and Sheng, Ying and Zheng, Lianmin and Yu, Cody Hao and Gonzalez, Joseph E. and Zhang, Hao and Stoica, Ion},
  booktitle = {Proceedings of the 29th Symposium on Operating Systems Principles},
  year = {2023},
  url = {https://arxiv.org/abs/2309.06180}
}

@article{lambert2024tulu3,
  title = {{Tulu 3: Pushing Frontiers in Open Language Model Post-Training}},
  author = {Lambert, Nathan and Morrison, Jacob and Pyatkin, Valentina and Huang, Shengyi and Ivison, Hamish and Brahman, Faeze and Miranda, Lester James V. and Liu, Alisa and Dziri, Nouha and Lyu, Shane and Gu, Yuling and Malik, Saumya and Graf, Victoria and Hwang, Jena D. and Yang, Jiangjiang and Bras, Ronan Le and Tafjord, Oyvind and Wilhelm, Chris and Soldaini, Luca and Smith, Noah A. and Wang, Yizhong and Dasigi, Pradeep and Hajishirzi, Hannaneh},
  journal = {arXiv preprint arXiv:2411.15124},
  year = {2024},
  url = {https://arxiv.org/abs/2411.15124}
}

@article{lin2026longtracerl,
  title = {{LongTraceRL: Learning Long-Context Reasoning from Search Agent Trajectories with Rubric Rewards}},
  author = {Lin, Nianyi and Zhang, Jiajie and Hou, Lei and Li, Juanzi},
  journal = {arXiv preprint arXiv:2605.31584},
  year = {2026},
  url = {https://arxiv.org/abs/2605.31584}
}

@article{ling2025longreason,
  title = {{LongReason: A Synthetic Long-Context Reasoning Benchmark via Context Expansion}},
  author = {Ling, Zhan and Liu, Kang and Yan, Kai and Yang, Yifan and Lin, Weijian and Fan, Ting-Han and Shen, Lingfeng and Du, Zhengyin and Chen, Jiecao},
  journal = {arXiv preprint arXiv:2501.15089},
  year = {2025},
  url = {https://arxiv.org/abs/2501.15089}
}

@article{liu2024bestpractices,
  title = {{Best Practices and Lessons Learned on Synthetic Data}},
  author = {Liu, Ruibo and Wei, Jerry and Liu, Fangyu and Si, Chenglei and Zhang, Yanzhe and Rao, Jinmeng and Zheng, Steven and Peng, Daiyi and Yang, Diyi and Zhou, Denny and Dai, Andrew M.},
  journal = {arXiv preprint arXiv:2404.07503},
  year = {2024},
  url = {https://arxiv.org/abs/2404.07503}
}

@article{liu2024lost,
  title = {{Lost in the Middle: How Language Models Use Long Contexts}},
  author = {Liu, Nelson F. and Lin, Kevin and Hewitt, John and Paranjape, Ashwin and Bevilacqua, Michele and Petroni, Fabio and Liang, Percy},
  journal = {Transactions of the Association for Computational Linguistics},
  year = {2024},
  url = {https://arxiv.org/abs/2307.03172}
}

@article{liu2025spiral,
  title = {{SPIRAL: Self-Play on Zero-Sum Games Incentivizes Reasoning via Multi-Agent Multi-Turn Reinforcement Learning}},
  author = {Liu, Bo and Guertler, Leon and Yu, Simon and Liu, Zichen and Qi, Penghui and Balcells, Daniel and Liu, Mickel and Tan, Cheston and Shi, Weiyan and Lin, Min and Lee, Wee Sun and Jaques, Natasha},
  journal = {arXiv preprint arXiv:2506.24119},
  year = {2025},
  url = {https://arxiv.org/abs/2506.24119}
}

@inproceedings{longpre2021entity,
  title = {{Entity-Based Knowledge Conflicts in Question Answering}},
  author = {Longpre, Shayne and Perisetla, Kartik and Chen, Anthony and Ramesh, Nikhil and DuBois, Chris and Singh, Sameer},
  booktitle = {Proceedings of the 2021 Conference on Empirical Methods in Natural Language Processing},
  year = {2021},
  url = {https://arxiv.org/abs/2109.05052}
}

@article{lv2026golongrl,
  title = {{GoLongRL: Capability-Oriented Long Context Reinforcement Learning with Multitask Alignment}},
  author = {Lv, Minxuan and Mei, Tiehua and Du, Tanlong and Chen, Junmin and Su, Zhenpeng and Chen, Ziyang and Wang, Ziqi and Wu, Zhennan and Pan, Ruotong and Liang, jian and Tang, Ruiming and Li, Han},
  journal = {arXiv preprint arXiv:2605.19577},
  year = {2026},
  url = {https://arxiv.org/abs/2605.19577}
}

@article{mukherjee2023orca,
  title = {{Orca: Progressive Learning from Complex Explanation Traces of GPT-4}},
  author = {Mukherjee, Subhabrata and Mitra, Arindam and Jawahar, Ganesh and Agarwal, Sahaj and Palangi, Hamid and Awadallah, Ahmed},
  journal = {arXiv preprint arXiv:2306.02707},
  year = {2023},
  url = {https://arxiv.org/abs/2306.02707}
}

@misc{openai2025mrcr,
  title = {{{OpenAI-MRCR}: Long context multiple needle in a haystack benchmark}},
  author = {{OpenAI}},
  howpublished = {\url{https://huggingface.co/datasets/openai/mrcr}},
  year = {2025}
}

@inproceedings{peng2024yarn,
  title = {{YaRN: Efficient Context Window Extension of Large Language Models}},
  author = {Peng, Bowen and Quesnelle, Jeffrey and Fan, Honglu and Shippole, Enrico},
  booktitle = {The Twelfth International Conference on Learning Representations},
  year = {2024},
  url = {https://arxiv.org/abs/2309.00071}
}

@inproceedings{sainz2023nlp,
  title = {{NLP Evaluation in trouble: On the Need to Measure LLM Data Contamination for each Benchmark}},
  author = {Sainz, Oscar and Campos, Jon Ander and García-Ferrero, Iker and Etxaniz, Julen and de Lacalle, Oier Lopez and Agirre, Eneko},
  booktitle = {Findings of the Association for Computational Linguistics: EMNLP 2023},
  year = {2023},
  url = {https://arxiv.org/abs/2310.18018}
}

@article{schulman2017ppo,
  title = {{Proximal Policy Optimization Algorithms}},
  author = {Schulman, John and Wolski, Filip and Dhariwal, Prafulla and Radford, Alec and Klimov, Oleg},
  journal = {arXiv preprint arXiv:1707.06347},
  year = {2017},
  url = {https://arxiv.org/abs/1707.06347}
}

@misc{schulman2020kl,
  title = {{Approximating KL Divergence}},
  author = {Schulman, John},
  howpublished = {\url{http://joschu.net/blog/kl-approx.html}},
  year = {2020}
}

@article{shao2024deepseekmath,
  title = {{DeepSeekMath: Pushing the Limits of Mathematical Reasoning in Open Language Models}},
  author = {Shao, Zhihong and Wang, Peiyi and Zhu, Qihao and Xu, Runxin and Song, Junxiao and Bi, Xiao and Zhang, Haowei and Zhang, Mingchuan and Li, Y. K. and Wu, Y. and Guo, Daya},
  journal = {arXiv preprint arXiv:2402.03300},
  year = {2024},
  url = {https://arxiv.org/abs/2402.03300}
}

@article{shen2025qwenlongl15,
  title = {{QwenLong-L1.5: Post-Training Recipe for Long-Context Reasoning and Memory Management}},
  author = {Shen, Weizhou and Yang, Ziyi and Li, Chenliang and Lu, Zhiyuan and Peng, Miao and Sun, Huashan and Shi, Yingcheng and Liao, Shengyi and Lai, Shaopeng and Zhang, Bo and Liu, Dayiheng and Huang, Fei and Zhou, Jingren and Yan, Ming},
  journal = {arXiv preprint arXiv:2512.12967},
  year = {2025},
  url = {https://arxiv.org/abs/2512.12967}
}

@inproceedings{sheng2025hybridflow,
  title = {{HybridFlow: A Flexible and Efficient RLHF Framework}},
  author = {Sheng, Guangming and Zhang, Chi and Ye, Zilingfeng and Wu, Xibin and Zhang, Wang and Zhang, Ru and Peng, Yanghua and Lin, Haibin and Wu, Chuan},
  booktitle = {Proceedings of the Twentieth European Conference on Computer Systems},
  year = {2025},
  url = {https://arxiv.org/abs/2409.19256}
}

@article{silver2017alphagozero,
  title = {{Mastering the game of {Go} without human knowledge}},
  author = {Silver, David and Schrittwieser, Julian and Simonyan, Karen and Antonoglou, Ioannis and Huang, Aja and Guez, Arthur and Hubert, Thomas and Baker, Lucas and Lai, Matthew and Bolton, Adrian and Chen, Yutian and Lillicrap, Timothy and Hui, Fan and Sifre, Laurent and van den Driessche, George and Graepel, Thore and Hassabis, Demis},
  journal = {Nature},
  volume = {550},
  number = {7676},
  pages = {354--359},
  year = {2017},
  url = {https://doi.org/10.1038/nature24270}
}

@article{stojanovski2025reasoninggym,
  title = {{REASONING GYM: Reasoning Environments for Reinforcement Learning with Verifiable Rewards}},
  author = {Stojanovski, Zafir and Stanley, Oliver and Sharratt, Joe and Jones, Richard and Adefioye, Abdulhakeem and Kaddour, Jean and Köpf, Andreas},
  journal = {arXiv preprint arXiv:2505.24760},
  year = {2025},
  url = {https://arxiv.org/abs/2505.24760}
}

@article{trivedi2022musique,
  title = {{MuSiQue: Multihop Questions via Single-hop Question Composition}},
  author = {Trivedi, Harsh and Balasubramanian, Niranjan and Khot, Tushar and Sabharwal, Ashish},
  journal = {Transactions of the Association for Computational Linguistics},
  year = {2022},
  url = {https://arxiv.org/abs/2108.00573}
}

@article{vodrahalli2024michelangelo,
  title = {{Michelangelo: Long Context Evaluations Beyond Haystacks via Latent Structure Queries}},
  author = {Vodrahalli, Kiran and Ontanon, Santiago and Tripuraneni, Nilesh and Xu, Kelvin and Jain, Sanil and Shivanna, Rakesh and Hui, Jeffrey and Dikkala, Nishanth and Kazemi, Mehran and Fatemi, Bahare and Anil, Rohan and Dyer, Ethan and Shakeri, Siamak and Vij, Roopali and Mehta, Harsh and Ramasesh, Vinay and Le, Quoc and Chi, Ed and Lu, Yifeng and Firat, Orhan and Lazaridou, Angeliki and Lespiau, Jean-Baptiste and Attaluri, Nithya and Olszewska, Kate},
  journal = {arXiv preprint arXiv:2409.12640},
  year = {2024},
  url = {https://arxiv.org/abs/2409.12640}
}

@article{wan2025qwenlongl1,
  title = {{QwenLong-L1: Towards Long-Context Large Reasoning Models with Reinforcement Learning}},
  author = {Wan, Fanqi and Shen, Weizhou and Liao, Shengyi and Shi, Yingcheng and Li, Chenliang and Yang, Ziyi and Zhang, Ji and Huang, Fei and Zhou, Jingren and Yan, Ming},
  journal = {arXiv preprint arXiv:2505.17667},
  year = {2025},
  url = {https://arxiv.org/abs/2505.17667}
}

@inproceedings{wang2023selfinstruct,
  title = {{Self-Instruct: Aligning Language Models with Self-Generated Instructions}},
  author = {Wang, Yizhong and Kordi, Yeganeh and Mishra, Swaroop and Liu, Alisa and Smith, Noah A. and Khashabi, Daniel and Hajishirzi, Hannaneh},
  booktitle = {Proceedings of the 61st Annual Meeting of the Association for Computational Linguistics (Volume 1: Long Papers)},
  year = {2023},
  url = {https://arxiv.org/abs/2212.10560}
}

@article{wang2025loongrl,
  title = {{LoongRL: Reinforcement Learning for Advanced Reasoning over Long Contexts}},
  author = {Wang, Siyuan and Zhang, Gaokai and Zhang, Li Lyna and Shang, Ning and Yang, Fan and Chen, Dongyao and Yang, Mao},
  journal = {arXiv preprint arXiv:2510.19363},
  year = {2025},
  url = {https://arxiv.org/abs/2510.19363}
}

@article{wang2025ultrafineweb,
  title = {{Ultra-FineWeb: Efficient Data Filtering and Verification for High-Quality LLM Training Data}},
  author = {Wang, Yudong and Fu, Zixuan and Cai, Jie and Tang, Peijun and Lyu, Hongya and Fang, Yewei and Zheng, Zhi and Zhou, Jie and Zeng, Guoyang and Xiao, Chaojun and Han, Xu and Liu, Zhiyuan},
  journal = {arXiv preprint arXiv:2505.05427},
  year = {2025},
  url = {https://arxiv.org/abs/2505.05427}
}

@article{xia2025agent0,
  title = {{Agent0: Unleashing Self-Evolving Agents from Zero Data via Tool-Integrated Reasoning}},
  author = {Xia, Peng and Zeng, Kaide and Liu, Jiaqi and Qin, Can and Wu, Fang and Zhou, Yiyang and Xiong, Caiming and Yao, Huaxiu},
  journal = {arXiv preprint arXiv:2511.16043},
  year = {2025},
  url = {https://arxiv.org/abs/2511.16043}
}

@article{xie2025logicrl,
  title = {{Logic-RL: Unleashing LLM Reasoning with Rule-Based Reinforcement Learning}},
  author = {Xie, Tian and Gao, Zitian and Ren, Qingnan and Luo, Haoming and Hong, Yuqian and Dai, Bryan and Zhou, Joey and Qiu, Kai and Wu, Zhirong and Luo, Chong},
  journal = {arXiv preprint arXiv:2502.14768},
  year = {2025},
  url = {https://arxiv.org/abs/2502.14768}
}

@article{xiong2025artificial,
  title = {{From Artificial Needles to Real Haystacks: Improving Retrieval Capabilities in LLMs by Finetuning on Synthetic Data}},
  author = {Xiong, Zheyang and Papageorgiou, Vasilis and Lee, Kangwook and Papailiopoulos, Dimitris},
  journal = {arXiv preprint arXiv:2406.19292},
  year = {2024},
  url = {https://arxiv.org/abs/2406.19292}
}

@article{xu2026beyond,
  title = {{Beyond Reward Engineering: A Data Recipe for Long-Context Reinforcement Learning}},
  author = {Xu, Xiaoyue and Zhang, Sikui and Wang, Xiaorong and Han, Xu and Xiao, Chaojun},
  journal = {arXiv preprint arXiv:2606.18831},
  year = {2026},
  url = {https://arxiv.org/abs/2606.18831}
}

@inproceedings{yang2018hotpotqa,
  title = {{HotpotQA: A Dataset for Diverse, Explainable Multi-hop Question Answering}},
  author = {Yang, Zhilin and Qi, Peng and Zhang, Saizheng and Bengio, Yoshua and Cohen, William W. and Salakhutdinov, Ruslan and Manning, Christopher D.},
  booktitle = {Proceedings of the 2018 Conference on Empirical Methods in Natural Language Processing},
  year = {2018},
  url = {https://arxiv.org/abs/1809.09600}
}

@article{yang2025qwen3,
  title = {{Qwen3 Technical Report}},
  author = {Yang, An and Li, Anfeng and Yang, Baosong and Zhang, Beichen and Hui, Binyuan and Zheng, Bo and Yu, Bowen and Gao, Chang and Huang, Chengen and Lv, Chenxu and Zheng, Chujie and Liu, Dayiheng and Zhou, Fan and Huang, Fei and Hu, Feng and Ge, Hao and Wei, Haoran and Lin, Huan and Tang, Jialong and Yang, Jian and Tu, Jianhong and Zhang, Jianwei and Yang, Jianxin and Yang, Jiaxi and Zhou, Jing and Zhou, Jingren and Lin, Junyang and Dang, Kai and Bao, Keqin and Yang, Kexin and Yu, Le and Deng, Lianghao and Li, Mei and Xue, Mingfeng and Li, Mingze and Zhang, Pei and Wang, Peng and Zhu, Qin and Men, Rui and Gao, Ruize and Liu, Shixuan and Luo, Shuang and Li, Tianhao and Tang, Tianyi and Yin, Wenbiao and Ren, Xingzhang and Wang, Xinyu and Zhang, Xinyu and Ren, Xuancheng and Fan, Yang and Su, Yang and Zhang, Yichang and Zhang, Yinger and Wan, Yu and Liu, Yuqiong and Wang, Zekun and Cui, Zeyu and Zhang, Zhenru and Zhou, Zhipeng and Qiu, Zihan},
  journal = {arXiv preprint arXiv:2505.09388},
  year = {2025},
  url = {https://arxiv.org/abs/2505.09388}
}

@article{yang2025spell,
  title = {{SPELL: Self-Play Reinforcement Learning for Evolving Long-Context Language Models}},
  author = {Yang, Ziyi and Shen, Weizhou and Li, Chenliang and Chen, Ruijun and Wan, Fanqi and Yan, Ming and Quan, Xiaojun and Huang, Fei},
  journal = {arXiv preprint arXiv:2509.23863},
  year = {2025},
  url = {https://arxiv.org/abs/2509.23863}
}

@article{yu2025dapo,
  title = {{DAPO: An Open-Source LLM Reinforcement Learning System at Scale}},
  author = {Yu, Qiying and Zhang, Zheng and Zhu, Ruofei and Yuan, Yufeng and Zuo, Xiaochen and Yue, Yu and Dai, Weinan and Fan, Tiantian and Liu, Gaohong and Liu, Lingjun and Liu, Xin and Lin, Haibin and Lin, Zhiqi and Ma, Bole and Sheng, Guangming and Tong, Yuxuan and Zhang, Chi and Zhang, Mofan and Zhang, Wang and Zhu, Hang and Zhu, Jinhua and Chen, Jiaze and Chen, Jiangjie and Wang, Chengyi and Yu, Hongli and Song, Yuxuan and Wei, Xiangpeng and Zhou, Hao and Liu, Jingjing and Ma, Wei-Ying and Zhang, Ya-Qin and Yan, Lin and Qiao, Mu and Wu, Yonghui and Wang, Mingxuan},
  journal = {arXiv preprint arXiv:2503.14476},
  year = {2025},
  url = {https://arxiv.org/abs/2503.14476}
}

@article{yu2025memagent,
  title = {{MemAgent: Reshaping Long-Context LLM with Multi-Conv RL-based Memory Agent}},
  author = {Yu, Hongli and Chen, Tinghong and Feng, Jiangtao and Chen, Jiangjie and Dai, Weinan and Yu, Qiying and Zhang, Ya-Qin and Ma, Wei-Ying and Liu, Jingjing and Wang, Mingxuan and Zhou, Hao},
  journal = {arXiv preprint arXiv:2507.02259},
  year = {2025},
  url = {https://arxiv.org/abs/2507.02259}
}

@inproceedings{yuan2024selfrewarding,
  title = {{Self-Rewarding Language Models}},
  author = {Yuan, Weizhe and Pang, Richard Yuanzhe and Cho, Kyunghyun and Li, Xian and Sukhbaatar, Sainbayar and Xu, Jing and Weston, Jason},
  booktitle = {Proceedings of the 41st International Conference on Machine Learning},
  year = {2024},
  url = {https://arxiv.org/abs/2401.10020}
}

@article{yue2026drzero,
  title = {{Dr. Zero: Self-Evolving Search Agents without Training Data}},
  author = {Yue, Zhenrui and Upasani, Kartikeya and Yang, Xianjun and Ge, Suyu and Nie, Shaoliang and Mao, Yuning and Liu, Zhe and Wang, Dong},
  journal = {arXiv preprint arXiv:2601.07055},
  year = {2026},
  url = {https://arxiv.org/abs/2601.07055}
}

@inproceedings{zelikman2022star,
  title = {{STaR: Bootstrapping Reasoning With Reasoning}},
  author = {Zelikman, Eric and Wu, Yuhuai and Mu, Jesse and Goodman, Noah D.},
  booktitle = {Advances in Neural Information Processing Systems},
  year = {2022},
  url = {https://arxiv.org/abs/2203.14465}
}

@article{zhao2025absolute,
  title = {{Absolute Zero: Reinforced Self-play Reasoning with Zero Data}},
  author = {Zhao, Andrew and Wu, Yiran and Yue, Yang and Wu, Tong and Xu, Quentin and Yue, Yang and Lin, Matthieu and Wang, Shenzhi and Wu, Qingyun and Zheng, Zilong and Huang, Gao},
  journal = {arXiv preprint arXiv:2505.03335},
  year = {2025},
  url = {https://arxiv.org/abs/2505.03335}
}

@article{zhao2025intuitor,
  title = {{Learning to Reason without External Rewards}},
  author = {Zhao, Xuandong and Kang, Zhewei and Feng, Aosong and Levine, Sergey and Song, Dawn},
  journal = {arXiv preprint arXiv:2505.19590},
  year = {2025},
  url = {https://arxiv.org/abs/2505.19590}
}

@article{zuo2025ttrl,
  title = {{TTRL: Test-Time Reinforcement Learning}},
  author = {Zuo, Yuxin and Zhang, Kaiyan and Sheng, Li and Qu, Shang and Cui, Ganqu and Zhu, Xuekai and Li, Haozhan and Zhang, Yuchen and Long, Xinwei and Hua, Ermo and Qi, Biqing and Sun, Youbang and Ma, Zhiyuan and Yuan, Lifan and Ding, Ning and Zhou, Bowen},
  journal = {arXiv preprint arXiv:2504.16084},
  year = {2025},
  url = {https://arxiv.org/abs/2504.16084}
}

\clearpage
\appendix
\raggedbottom

\section{Implementation Details}
\label{sec:details}

\subsection{Experiment Details}
\label{sec:exp-details}

\paragraph{Seed lists.}
Reasoning worlds are written in English in one of 12 domains (tech industry, biotech/pharma, maritime shipping, aerospace suppliers, mining \& energy, football clubs, semiconductor equipment, satellite communications, rail freight operators, wind \& tidal energy, hotel \& resort chains, specialty chemicals). Each world samples its document genres from 8 (magazine feature, corporate history chapter, analyst research note, conference proceedings digest, podcast interview transcript, obituary, regulatory filing summary, trade-press news roundup). The style card fixes a 40-year period whose start is drawn from the decades 1870--1990, one of 12 naming styles (Nordic, Iberian, Slavic, Dutch/Flemish, Scottish, Franco-Belgian, Greek, Turkish, Hungarian, Baltic, Irish, Italian), and a list of overused names that must not appear (Appendix~\ref{sec:prompts}).

\begin{table}[!h]
\centering
\resizebox{\columnwidth}{!}{%
\begin{tabular}{llr}
\toprule
\textbf{Stage} & \textbf{Decoding} & \textbf{Max tokens} \\
\midrule
Specification & thinking, $T{=}0.7$/$0.6$ & 28{,}672 \\
Plan & $T{=}0.8$ (non-thinking$^\dagger$) & 16{,}384 \\
Expansion & $T{=}0.8$ (non-thinking$^\dagger$) & 5{,}120 \\
Difficulty filter & thinking, $T{=}0.7$ & 8{,}192 \\
\bottomrule
\end{tabular}%
}
\caption{Generation settings for reasoning data, all with top-$p$ 0.95. $T{=}0.7$ applies to Qwen3-4B-Thinking-2507, which always thinks, and $T{=}0.6$ to Qwen3-8B with thinking. $^\dagger$Qwen3-8B only.}
\label{tab:gen-params}
\end{table}

\begin{table}[!h]
\centering
\resizebox{\columnwidth}{!}{%
\begin{tabular}{lr}
\toprule
\textbf{Hyperparameter} & \textbf{Value} \\
\midrule
Algorithm & GRPO (verl) \\
Batch per step & 64 prompts $\times$ 8 \\
Training steps & 72 (4B), 108 (8B) \\
PPO update & 16 prompts, 1 epoch \\
Optimizer & AdamW (0.9, 0.999) \\
Learning rate & $10^{-6}$, constant \\
Weight decay, grad.\ clip & 0.01, 1.0 \\
Clipping & $\varepsilon{=}0.2$, dual-clip $c{=}3$ \\
KL penalty & $\beta{=}0.001$, low-var., in loss \\
Entropy bonus & 0 \\
Loss aggregation & token mean \\
Max prompt length & 32{,}768 \\
Max response length & 8{,}192 \\
Rollout sampling & $T{=}1.0$, top-$p{=}1.0$ \\
Ulysses parallelism & 2 \\
Answer-length tolerance & 128 characters \\
\bottomrule
\end{tabular}%
}
\caption{RL hyperparameters shared by all methods.}
\label{tab:rl-params}
\end{table}

\paragraph{Reasoning data generation.}
All generation calls use the model being trained, served with vLLM \citep{kwon2023vllm}.
Qwen3-4B-Thinking-2507 always reasons before answering; we treat each call's token budget as the total for reasoning and answer, and reject outputs without a closing \texttt{</think>}.
For Qwen3-8B, thinking is enabled for the specification and disabled for planning and expansion.
Table~\ref{tab:gen-params} lists the per-stage settings.

The specification prompt asks for 8--14 entities, $Q{=}4$ question chains of $H_q{=}3$--5 hops, and $K_q{=}2$--3 distractor facts per chain, and each world has $M{=}8$ documents.
The specification and the plan are each attempted at most twice; a world is rejected if either fails to parse. For Qwen3-8B, a world is also rejected if its specification does not contain exactly four questions or its plan does not contain exactly eight documents.
A program keeps only questions whose supporting facts all exist, and keeps only aliases that are equivalent surface forms of the answer (added or removed articles and company suffixes, numbers written as words, or sub-spans of at least two content words); aliases that restate the question are removed.

The plan is repaired by removing hallucinated or duplicated fact IDs and placing any missing fact into the currently smallest document; each document lists up to 15 facts of other documents as forbidden.
Expansion targets 2{,}400--2{,}700 words per document for Qwen3-4B-Thinking-2507 and 4{,}000--4{,}500 words for Qwen3-8B, which writes shorter documents for the same target; documents longer than 14{,}000 characters are truncated at the last paragraph boundary, and a document that leaks instructions into the text (e.g., ``facts to embed'') is regenerated once and otherwise rejected.

Worlds whose estimated context exceeds 27.5K tokens are discarded. The faithfulness gate normalizes case, whitespace, and punctuation; bare-year answers are exempt from it, since years frequently appear in unrelated sentences (at most one question per world may have such an answer).
In the training prompt, the eight shuffled documents are concatenated without document markers.

\paragraph{Difficulty filtering.}
The filter uses a separate prompt that presents the eight documents with \texttt{<CHUNK\_i>} markers and asks for the evidence documents and a short answer in \texttt{<answer>} tags (Appendix~\ref{sec:prompts}); the answer is matched with the same whole-word rule as the reasoning reward. Each wave draws two samples (temperature 0.7, top-$p$ 0.95), and a question receives at most $W{=}4$ answers; requests that fail at the server are not counted as answers. The validation split is defined by a hash of the world identifier, so that no world appears in both splits.

\paragraph{Retrieval data generation.}
The 10 formats are diary entry, email, formal letter, poem, riddle, short essay, short news article, short scene in a play, social media post, and song; the 434 nouns are those of the OpenAI MRCR dataset. The model answers each of the 4{,}340 requests $R{=}100$ times, giving a pool of 434K responses.

Responses are sampled at temperature 0.95. For Qwen3-8B, they are generated through the chat template with thinking disabled and at most 1{,}024 tokens. Qwen3-4B-Thinking-2507 cannot disable thinking: when the prompt is prefilled with an empty reasoning block, it still writes reasoning followed by \texttt{</think>}. We therefore allow 3{,}072 tokens, keep the text after the last \texttt{</think>}, and resample (up to three attempts) when no \texttt{</think>} is produced.

To assemble a conversation, the program draws a target (format, noun) pair and $N{=}8$ distinct pooled responses to it, draws a target length uniformly from 8{,}193--16{,}384 tokens (counted with the generating model's chat template), and adds distractor turns, requests for other pairs with pooled responses, until the length is reached; the needles are inserted at uniformly random positions.

The first user turn contains two worked examples of the task, and the final turn asks for the $k$-th needle, $k$ uniform in $\{1,\dots,8\}$, prefixed by a random 10-character alphanumeric string (Appendix~\ref{sec:prompts}). Conversations outside the length range are re-drawn. The median length is 11.4K (p10--p90: 9.0K--14.5K) tokens for Qwen3-4B-Thinking-2507 and 11.1K (8.8K--13.9K) for Qwen3-8B.

\paragraph{RL training.}
Table~\ref{tab:rl-params} lists the RL hyperparameters, which are shared by \LongZero and both baselines.
Runs on Qwen3-4B-Thinking-2507 take 72 steps and runs on Qwen3-8B take 108 steps; with 3{,}456 training samples these are about 1.3 and 2 epochs. The answer-length tolerance of the reasoning reward is $L{=}128$ characters.
Rollouts use vLLM with tensor parallelism 1 for Qwen3-4B-Thinking-2507 and 2 for Qwen3-8B, and training uses Ulysses sequence parallelism \citep{jacobs2023ulysses}.
We evaluate on the held-out validation set every 6 steps with greedy decoding and save a checkpoint every 18 steps.
We did not run a hyperparameter search for the final experiments: the configuration was fixed during earlier development runs on previous versions of our data and is shared by all methods, so no hyperparameter was tuned on the evaluation benchmarks.

\paragraph{Baselines.}
\textbf{Enhanced HotpotQA (from LoongRL).} We use the distractor-augmented HotpotQA data produced by LoongRL's data pipeline, without the UUID key chains of its KeyChain variant. To reduce cost, difficulty filtering uses Qwen3-4B-Instruct-2507 instead of the 32B model of the original pipeline and does not use a thinking model; questions answered correctly 1--6 times out of 8 are kept.

The 5{,}120 training questions (median context 14.4K tokens) use LoongRL's prompt, which asks the model to end with ``\texttt{Answer:}'' followed by the answer in \texttt{\textbackslash boxed\{\}}. The reward follows LoongRL's exact-match reward: the response must contain exactly one \texttt{\textbackslash boxed\{\}}, the normalized gold answer must be contained in its content, the content may exceed the gold answer by at most 128 characters, and at most 32 characters may follow the boxed answer. Validation uses 128 questions randomly sampled from LoongRL's validation split. Checkpoints are saved every 36 steps. The same data are used for both base models.

\textbf{Teacher-synthesized QA.} For each document cluster, Qwen3.5-397B-A17B proposes a question and answer, then answers its own question up to five times and judges semantic equivalence with its reference answer; it generates up to five rounds of increasingly difficult questions per cluster, of which we use only the first round (4{,}291 questions).

For filtering, each base model answers every question eight times with its default thinking mode (temperature 0.6, top-$p$ 0.95, top-$k$ 20, at most 8{,}192 tokens), and Qwen3-4B-Thinking-2507 judges semantic equivalence with the gold answer (temperature 0, numerical tolerance 1\%); the answering and verification prompts are those released with SPELL. Questions answered correctly 1--7 times are eligible; after removing duplicated documents, 2{,}657 (Qwen3-4B-Thinking-2507) and 2{,}644 (Qwen3-8B) remain, from which 2{,}304 training and 256 validation questions with disjoint documents are sampled. Their median prompt is about 4.7K tokens (maximum about 15K). Training uses the same prompt template and reasoning reward as \LongZero, with the teacher's answer as the only accepted answer.

\paragraph{Evaluation.}
All benchmarks are served with vLLM; Qwen3-4B-Thinking-2507 uses a 262{,}144-token window and Qwen3-8B YaRN (factor 4, 131{,}072 tokens).
Unless stated otherwise, the maximum output length is 16{,}384 tokens, the prediction is the text after the final \texttt{</think>}, and a response without \texttt{</think>} is resampled up to two times before being counted as invalid.
\textbf{LB1}: official prompts and F1 (600 questions), without truncating contexts.
\textbf{LBv2}: the official zero-shot prediction script with up to 51{,}200 output tokens; inputs are truncated in the middle to 120K tokens (Qwen3-4B-Thinking-2507) or 79K tokens (Qwen3-8B).
\textbf{FRAMES}: all oracle articles are placed in the context; for Qwen3-8B the article text is limited to about 380K characters to fit the window. A prediction is correct if the normalized answer contains, or is contained in, the normalized gold answer.
\textbf{AA-LCR}: documents are concatenated in the order given by the dataset and used with the official prompt; for Qwen3-8B the output budget is the window minus the input length. Answers are judged by GPT-5.6-Luna (medium reasoning effort) with the official AA-LCR v1.1 judging prompts; API failures are retried and never counted as incorrect answers.
\textbf{LongReason}: 32K split with the prompts provided by the dataset; the predicted option is the first letter A--E after the last ``the answer is''.

\paragraph{Compute budget.}
RL training runs on a single node with 8 NVIDIA A800-SXM4-80GB GPUs. Training \LongZero takes about 27 hours on Qwen3-4B-Thinking-2507 and 29 hours on Qwen3-8B, that is, about 213 and 230 GPU-hours.
Evaluation runs on nodes with 8 NVIDIA RTX 5090 or 4090 GPUs; evaluating one checkpoint on all five benchmarks with two runs each takes about 12 hours, roughly 100 GPU-hours.
AA-LCR answers are judged by GPT-5.6-Luna through its API with medium reasoning effort, which takes 200 judge calls per evaluated checkpoint.

\subsection{Prompt Templates}
\label{sec:prompts}

We list the prompts of LongZero. Placeholders are shown in braces.

\subsubsection*{\LongZero: reasoning data}

\begin{promptbox}{World specification}
You are constructing a synthetic "world" for long-context QA training data. ALL entities must be fictional (realistic style, but names/companies/events must not correspond to real ones). World language: {lang_name}. Domain: {domain}.

Style card (mandatory):
{style_card}

Design requirements (all mandatory):
1. 8-14 named entities. At least two PEOPLE share a surname, and at least two ORGANIZATIONS have confusably similar names.
2. Create 4 QUESTION CHAINS of 3-5 hops each. A hop = one atomic fact linking two entities/attributes. Chains must partially share bridge entities.
3. For each chain, create 2-3 DISTRACTOR facts that share the bridge entity but differ in role/time/attribute (e.g. founder vs CEO, predecessor product, earlier residence, similarly-named org).
4. Every fact gets an id F1..Fn (distractors D1..Dn). Questions reference supporting fact ids.
5. In the QUESTION text, refer to intermediate/bridge entities descriptively rather than by name (e.g. "the founder of the magazine's parent press"). The ANSWER itself must be the concrete name, place, product, title or number exactly as it appears in a fact — NEVER a description such as "the person who founded X" or "the lead engineer of Y".
6. At most ONE question may have a bare year as its answer; prefer names, places, products, titles.

Output STRICT JSON only, no markdown fences, exactly this schema:
{"entities": [{"name": "...", "type": "person|org|place|product", "note": "..."}],
 "facts": [{"id": "F1", "text": "...", "entities": ["..."]}],
 "distractors": [{"id": "D1", "text": "...", "confuses_with": "F3", "type": "role|time|name|place"}],
 "questions": [{"qid": "Q1", "question": "...", "answer": "...", "aliases": ["..."],
   "supporting_facts": ["F1","F4","F7"], "n_hops": 4}]}
\end{promptbox}

\begin{promptbox}{Style card (inserted as \{style\_card\})}
- Set the story mainly in the {era}s-{era+40}s.
- Person and company names should have a {flavor} flavor (fictional, varied, not stereotypical).
- NEVER use any of these overused names or similar variants: Voss, Carter, Smith, Vane, Vance, Hart, Lee, Elara, Marcus, James, NexTech, QuantumCore, TechNova, MarineTech, AeroCore, AeroTech, <common names>.
\end{promptbox}

\begin{promptbox}{Plan}
You will distribute facts into {n_chunks} documents for a long-context QA task. World language: {lang_name}. Available document genres: {genres}.

Facts (must ALL be placed, each in exactly one document):
{facts}

Rules:
- Facts supporting the same question should be spread across DIFFERENT documents where possible.
- Each document gets a genre and a one-line topic.
- Distractor facts go in documents DIFFERENT from the fact they confuse with, where possible.

Output STRICT JSON only:
{"chunks": [{"chunk_id": 1, "genre": "...", "topic": "...", "fact_ids": ["F1","D2"]}]}
\end{promptbox}

\begin{promptbox}{Expansion (one call per document)}
Write a document in {lang_name}. Genre: {genre}. Topic: {topic}. Target length: {target_words}.

The document MUST naturally state each of the following facts exactly once, without altering any name, number, or relationship.

Facts to embed:
{facts}

Hard constraints:
- Do NOT mention, quote, or allude to these instructions, or your embedding strategy anywhere in the document. No meta-commentary about how facts are placed. Pure in-world prose only.
- Do NOT state any of these forbidden facts, even in passing: {forbidden}
- Do NOT summarize or foreshadow other documents.
- Match the genre's register convincingly; include realistic filler content of your own invention that does not contradict the facts.
- Output the document text only, no titles about the task, no explanations.
\end{promptbox}

\begin{promptbox}{Difficulty filter}
You are given several documents, each wrapped in <CHUNK_i> tags. Answer the question using ONLY information from the documents.
{context}
Question: {question}

First identify which chunks contain the evidence needed, then give a short final answer. Respond in exactly this format:
<useful_chunks> CHUNK_i, CHUNK_j, ... </useful_chunks>
<answer> your short answer </answer>
\end{promptbox}

\begin{promptbox}{RL training on reasoning data (also used for the teacher-synthesized baseline)}
Answer the question based on the given passages.

Reason step by step, making sure every step is backed by the passages. End your response with a final line: The answer is <answer>. Keep the answer precise and succinct.

The following are the given passages.
{documents}

Question: {question}
\end{promptbox}

\subsubsection*{\LongZero: retrieval data}

\begin{promptbox}{Response pool (single-turn request)}
write a {format} about {noun}
\end{promptbox}

\begin{promptbox}{Conversation template (multi-turn; the first user turn contains two example blocks)}
[user]
Here are some examples of conversations succeeded by a follow-up question answered correctly:

======EXAMPLE======
User: write a {format1} about {noun1}
Assistant: {response1_1}
User: write a {format2} about {noun2}
Assistant: {response2}
User: write a {format1} about {noun1}
Assistant: {response1_2}
User: Prepend {s'} to the 1st (1 indexed) {format1} about {noun1}. Do not include any other text in your response.
Assistant: {s'}{response1_1}
======END EXAMPLE======

======EXAMPLE======
(a second example block of the same form)
======END EXAMPLE======
[user]       write a {format_j} about {noun_j}
[assistant]  {pooled response}
...          (distractor turns and the 8 needle turns "write a {format} about {noun}")
[user]
Prepend {s} to the {ordinal} (1 indexed) {format} about {noun}. Do not include any other text in your response.
\end{promptbox}
Here \texttt{\{ordinal\}} is 1st, 2nd, 3rd, or $k$th, and \texttt{\{s\}} is a random 10-character alphanumeric string.

\subsubsection*{Baselines and evaluation}

Public prompts are not repeated: the baselines use the prompts of LoongRL \citep{wang2025loongrl} and SPELL \citep{yang2025spell}, and the benchmarks use their official prompts, except FRAMES, for which we list the articles under ``\texttt{\# Article: \{title\}}'' headings before the question.

\newpage
\section{Additional Results}
\label{sec:additional}

Tables~\ref{tab:lb1-4b}--\ref{tab:lbv2-8b} break down the results of Table~\ref{tab:main}, and Tables~\ref{tab:lb1-abl} and~\ref{tab:lbv2-abl} those of Table~\ref{tab:ablation}, by dataset for LB1, namely HotpotQA (HQA), 2WikiMultihopQA (2WQA), and MuSiQue (MSQ), and by difficulty and input length for LBv2, for Qwen3-4B-Thinking-2507 and Qwen3-8B. As in Table~\ref{tab:main}, the best result in each column is in \textbf{bold} and the second best is \underline{underlined}.

\begin{table}[!h]
\centering
\setlength{\tabcolsep}{4pt}
\resizebox{\columnwidth}{!}{%
\begin{NiceTabular}{lcccc}
\CodeBefore
\columncolor{avgbg}{5}
\Body
\toprule
\textbf{Method} & \textbf{HQA} & \textbf{2WQA} & \textbf{MSQ} & \avg{\textbf{Avg}} \\
\midrule
Base model & 67.04 & \snd{78.46} & 50.79 & \avg{65.43} \\
Enhanced HotpotQA & 66.59 & 76.41 & \snd{54.82} & \avg{65.94} \\
Teacher-synthesized QA & \fst{67.73} & 77.97 & 52.68 & \avg{\snd{66.12}} \\
\LongZero (Ours) & \snd{67.17} & \fst{79.62} & \fst{54.86} & \avg{\fst{67.22}} \\
\bottomrule
\end{NiceTabular}%
}
\caption{LB1 F1 of Qwen3-4B-Thinking-2507 on each of its three datasets. Avg is the LB1 score in Table~\ref{tab:main}.}
\label{tab:lb1-4b}
\end{table}

\begin{table}[!h]
\centering
\setlength{\tabcolsep}{4pt}
\resizebox{\columnwidth}{!}{%
\begin{NiceTabular}{lcccc}
\CodeBefore
\columncolor{avgbg}{5}
\Body
\toprule
\textbf{Method} & \textbf{HQA} & \textbf{2WQA} & \textbf{MSQ} & \avg{\textbf{Avg}} \\
\midrule
Base model & 66.58 & \snd{76.83} & 47.48 & \avg{63.63} \\
Enhanced HotpotQA & \fst{67.91} & 75.43 & \fst{51.60} & \avg{\snd{64.98}} \\
Teacher-synthesized QA & \snd{66.99} & 76.12 & 49.08 & \avg{64.07} \\
\LongZero (Ours) & 66.90 & \fst{78.20} & \snd{51.38} & \avg{\fst{65.49}} \\
\bottomrule
\end{NiceTabular}%
}
\caption{LB1 F1 of Qwen3-8B on each of its three datasets, which contain 200 questions each. Avg is the LB1 score in Table~\ref{tab:main}.}
\label{tab:lb1-8b}
\end{table}

\begin{table}[!h]
\centering
\setlength{\tabcolsep}{4pt}
\resizebox{\columnwidth}{!}{%
\begin{NiceTabular}{lcccccc}
\CodeBefore
\columncolor{avgbg}{7}
\Body
\toprule
\Block{2-1}{\textbf{Method}} & \multicolumn{2}{c}{\textbf{Difficulty}} & \multicolumn{3}{c}{\textbf{Length}} & \Block{2-1}{\textbf{Overall}} \\
\cmidrule(lr){2-3}\cmidrule(lr){4-6}
& \textbf{Easy} & \textbf{Hard} & \textbf{Short} & \textbf{Medium} & \textbf{Long} & \\
\midrule
Base model & \snd{48.44} & 35.69 & 47.78 & \snd{37.44} & 34.72 & \avg{40.56} \\
\Block[l]{}{Enhanced\\HotpotQA} & \fst{49.74} & \snd{38.91} & 51.67 & \fst{38.37} & \snd{37.96} & \avg{\snd{43.04}} \\
\Block[l]{}{Teacher-\\synthesized QA} & 47.66 & 38.26 & \snd{51.94} & 35.12 & \fst{38.43} & \avg{41.85} \\
\LongZero (Ours) & 46.88 & \fst{41.16} & \fst{53.89} & \fst{38.37} & 35.65 & \avg{\fst{43.34}} \\
\bottomrule
\end{NiceTabular}%
}
\caption{LBv2 accuracy of Qwen3-4B-Thinking-2507 by difficulty and by input length, using the labels provided with LBv2. Overall is the LBv2 score in Table~\ref{tab:main}.}
\label{tab:lbv2-4b}
\end{table}

\begin{table}[!h]
\centering
\setlength{\tabcolsep}{4pt}
\resizebox{\columnwidth}{!}{%
\begin{NiceTabular}{lcccccc}
\CodeBefore
\columncolor{avgbg}{7}
\Body
\toprule
\Block{2-1}{\textbf{Method}} & \multicolumn{2}{c}{\textbf{Difficulty}} & \multicolumn{3}{c}{\textbf{Length}} & \Block{2-1}{\textbf{Overall}} \\
\cmidrule(lr){2-3}\cmidrule(lr){4-6}
& \textbf{Easy} & \textbf{Hard} & \textbf{Short} & \textbf{Medium} & \textbf{Long} & \\
\midrule
Base model & \fst{40.89} & 26.69 & \snd{45.00} & \snd{25.35} & 24.07 & \avg{\snd{32.11}} \\
\Block[l]{}{Enhanced\\HotpotQA} & 37.24 & \snd{27.65} & 44.17 & 23.49 & \snd{25.46} & \avg{31.31} \\
\Block[l]{}{Teacher-\\synthesized QA} & \snd{38.54} & \fst{29.90} & \fst{45.83} & \fst{27.21} & 24.07 & \avg{\fst{33.20}} \\
\LongZero (Ours) & \fst{40.89} & 26.53 & 43.89 & 25.12 & \fst{25.93} & \avg{32.01} \\
\bottomrule
\end{NiceTabular}%
}
\caption{LBv2 accuracy of Qwen3-8B by difficulty and by input length, using the labels provided with LBv2. Overall is the LBv2 score in Table~\ref{tab:main}.}
\label{tab:lbv2-8b}
\end{table}

\makeatletter\setlength{\@fptop}{0pt}\setlength{\@fpsep}{12pt plus 2pt}\setlength{\@fpbot}{0pt plus 1fil}\makeatother
\begin{table}[t]
\centering
\setlength{\tabcolsep}{4pt}
\resizebox{\columnwidth}{!}{%
\begin{NiceTabular}{lcccc}
\CodeBefore
\columncolor{avgbg}{5}
\Body
\toprule
\textbf{Training data} & \textbf{HQA} & \textbf{2WQA} & \textbf{MSQ} & \textbf{Avg} \\
\midrule
Base model & 67.04 & 78.46 & 50.79 & 65.43 \\
Reasoning data only & \fst{68.51} & 78.45 & 54.15 & 67.04 \\
Retrieval data only & 64.57 & 78.48 & 50.48 & 64.51 \\
\Block[l]{}{Reasoning and\\retrieval data, 1:1} & \snd{67.33} & \snd{79.58} & \fst{54.96} & \fst{67.29} \\
\Block[l]{}{Reasoning and retrieval\\data, 2:1 (Ours)} & 67.17 & \fst{79.62} & \snd{54.86} & \snd{67.22} \\
\bottomrule
\end{NiceTabular}%
}
\caption{LB1 F1 of the ablation variants of Qwen3-4B-Thinking-2507 on each dataset. Avg is the LB1 score in Table~\ref{tab:ablation}.}
\label{tab:lb1-abl}
\end{table}

\begin{table}[t]
\centering
\setlength{\tabcolsep}{4pt}
\resizebox{\columnwidth}{!}{%
\begin{NiceTabular}{lcccccc}
\CodeBefore
\columncolor{avgbg}{7}
\Body
\toprule
\Block{2-1}{\textbf{Training}\\\textbf{data}} & \multicolumn{2}{c}{\textbf{Difficulty}} & \multicolumn{3}{c}{\textbf{Length}} & \Block{2-1}{\textbf{Overall}} \\
\cmidrule(lr){2-3}\cmidrule(lr){4-6}
 & \textbf{Easy} & \textbf{Hard} & \textbf{Short} & \textbf{Medium} & \textbf{Long} & \\
\midrule
\Block[l]{}{Base\\model} & \snd{48.44} & 35.69 & 47.78 & 37.44 & 34.72 & 40.56 \\
\Block[l]{}{Reasoning\\data only} & \fst{51.04} & 37.46 & 48.33 & \fst{39.77} & \fst{38.89} & \snd{42.64} \\
\Block[l]{}{Retrieval\\data only} & 44.27 & \snd{40.03} & \snd{53.06} & 36.51 & 32.87 & 41.65 \\
\Block[l]{}{Reasoning and\\retrieval, 1:1} & 46.88 & 37.94 & 49.44 & 36.05 & \snd{38.43} & 41.35 \\
\Block[l]{}{Reasoning and\\retrieval, 2:1\\(Ours)} & 46.88 & \fst{41.16} & \fst{53.89} & \snd{38.37} & 35.65 & \fst{43.34} \\
\bottomrule
\end{NiceTabular}%
}
\caption{LBv2 accuracy of the ablation variants of Qwen3-4B-Thinking-2507 by difficulty and by input length. Overall is the LBv2 score in Table~\ref{tab:ablation}.}
\label{tab:lbv2-abl}
\end{table}

\clearpage
\section{Case Studies}
\label{sec:cases}
We present three types of questions: a fully correct question that is kept for training, a defective question that difficulty filtering removes, and a defective question that passes difficulty filtering and enters the training set.

\newtcolorbox{casebox}[1]{enhanced, colback=white, colframe=black!45, boxrule=0.5pt, arc=1pt,
  left=4pt, right=4pt, top=3pt, bottom=3pt, fonttitle=\bfseries\small, coltitle=black, colbacktitle=pbg, title={#1}, fontupper=\small}

\subsection{A Fully Correct Question}
\begin{casebox}{Question W188\_Q1, kept for training}
\textbf{Question:} Who was the researcher who used the Vitae Extract developed by Aurora Biologica in 1885?\\
\textbf{Answer:} Sofia Bianchi.\\
\textbf{Supporting facts:} F2 Aurora Biologica developed Vitae Extract in 1885, in Document 1; F3 Sofia Bianchi used Vitae Extract in 1887, in Document 8.\\
\textbf{Distractors:} D2 Sofia Bianchi was the head of the research division at Aurora Biologica in 1886, in Document 1; D3 Sofia Bianchi worked on the development of the Luminosa Antiseptic for Aurora Farmaceutica in 1890, in Document 5.\\
\textbf{Document 1:} ``By the end of 1885, Vitae Extract had become Aurora Biologica's first commercially viable product \dots''\\
\textbf{Document 8:} ``Sofia Bianchi's work with Vitae Extract in 1887 provides a valuable case study for understanding the intersection of early research and regulatory oversight.''\\
\textbf{Filtering answers:} Sofia Bianchi \ding{51}, Marco Rossi \ding{55}.
\end{casebox}
The question is consistent with every document: the two supporting facts are stated in different documents, and the distractors give Sofia Bianchi other roles at other times without contradicting F3. The question therefore has exactly one correct answer. It is not trivial for the model either: one of the two filtering answers was Marco Rossi, the founder of Aurora Biologica, who is introduced in the same document as F2. Because the two answers disagree, the question falls in the unstable band and is kept.

\subsection{A Defective Question Removed by Filtering}
\begin{casebox}{Question W188\_Q4, discarded}
\textbf{Question:} In what year was Soleil Tincture first used by Enrico Vittori?\\
\textbf{Answer:} 1897, from F11 Soleil Tincture was used by Enrico Vittori in 1897.\\
\textbf{Distractors:} D8 Enrico Vittori used Soleil Tincture in the development of the Luminosa Antiseptic in 1890; D9 Soleil Tincture was used by Enrico Vittori in 1896.\\
\textbf{Documents:} ``Vittori's application of Soleil Tincture in the development process in 1890 was instrumental \dots''; ``Soleil Tincture's documented use by Enrico Vittori in 1896 and 1897 was primarily within the context of specific, localized outbreaks \dots''\\
\textbf{Filtering answers:} 1896 \ding{55}, 1896 \ding{55}, 1890 \ding{55}, 1890 \ding{55}.
\end{casebox}
The answer passes the faithfulness gate, because 1897 appears in the text, but the question is wrong: two distractors state earlier uses by the same person, so according to the documents the first use is not 1897. All four answers disagree with the reference answer, and the question is discarded. This case illustrates why we drop questions that the model never answers correctly; many of them are defective rather than hard.

\subsection{A Defective Question That Passes Filtering}
\begin{casebox}{Question W81\_Q1, kept for training}
\textbf{Question:} Who founded the rail freight operator that introduced the Aegean-1 rail car model and deployed it in Korinthos?\\
\textbf{Answer:} Demetrios Kostas, from F1 Demetrios Kostas founded Aegean Railways.\\
\textbf{Distractor:} D2 Anastasia Kostas founded Aegean Railways, written as a distractor for F1.\\
\textbf{Documents:} ``Demetrios Kostas's role as the founder and first CEO was pivotal in establishing this legacy \dots''; ``Anastasia Kostas founded Aegean Railways, creating a dedicated entity that would specialize in rail freight operations \dots''\\
\textbf{Filtering answers:} Anastasia Kostas \ding{55}, Demetrios Kostas \ding{51}.
\end{casebox}
The specification prompt asks for distractors that share an entity with a fact but differ in role, time, or attribute. D2 instead assigns the same role to a different person with a confusable name and thus contradicts F1, so the documents support two answers. Because the model answered once with each name, the question entered the unstable band and the training set, where the reward penalizes an answer that the context supports. Difficulty filtering cannot detect such contradictions whenever the model sometimes reproduces the reference answer; a consistency check of the specification before expansion would remove such cases.

\end{document}